%% file: main.tex
\documentclass[10pt,twocolumn,letterpaper]{article}

\usepackage[pagenumbers]{cvpr} % To force page numbers, e.g. for an arXiv version

\input{preamble}

\definecolor{cvprblue}{rgb}{0.21,0.49,0.74}
\usepackage[pagebackref,breaklinks,colorlinks,citecolor=cvprblue]{hyperref}

\def\paperID{10104} % *** Enter the Paper ID here
\def\confName{CVPR}
\def\confYear{2024}

\title{D$^4$M: Dataset Distillation via Disentangled Diffusion Model}

\author{
Duo Su$^{1,5,6,\dagger}$ \quad\quad
Junjie Hou$^{2,5,6,\dagger}$\quad\quad
Weizhi Gao$^3$\quad\quad
Yingjie Tian$^{4,5,6,7,*}$\quad\quad
Bowen Tang$^8$\vspace{0.5em}\\
$^1$School of Computer Science and Technology, UCAS \quad\quad   
$^2$Sino-Danish College, UCAS\\
$^3$Department of Computer Science, NCSU\quad\quad  
$^4$School of Economics and Management, UCAS\\
$^5$Research Center on Fictitious Economy and Data Science, CAS\\
$^6$Key Laboratory of Big Data Mining and Knowledge Management, CAS \\
$^7$MOE Social Science Laboratory of Digital Economic Forecasts and Policy Simulation, UCAS\\
$^8$Institute of Computing Technology, CAS \\
\url{https://junjie31.github.io/D4M/}
}

\begin{document}
\maketitle
\newcommand\blfootnote[1]{%
\begingroup
\renewcommand\thefootnote{}\footnote{#1}%
\addtocounter{footnote}{-1}%
\endgroup
}
\blfootnote{$\dagger$ Equal contribution. $*$ Corresponding author.} 
\input{sec/0_abstract}    
\input{sec/1_intro}
\input{sec/2_related}
\input{sec/3_method}

\input{sec/4_experiment}

\input{sec/5_conclusion}
{
    \small
    \bibliographystyle{ieeenat_fullname}
    \bibliography{main}
}
\input{sec/X_suppl}


\end{document}

%% file: preamble.tex
\usepackage[dvipsnames]{xcolor}

\usepackage{multirow}
\usepackage{colortbl}
\usepackage{amsthm,amsmath,amssymb}
\usepackage{algorithm}
\usepackage{algorithmic}

%% file: sec/0_abstract.tex
\begin{abstract}
Dataset distillation offers a lightweight synthetic dataset for fast network training with promising test accuracy. 
To imitate the performance of the original dataset, most approaches employ bi-level optimization and the distillation space relies on the matching architecture.
Nevertheless, these approaches either suffer significant computational costs on large-scale datasets or experience performance decline on cross-architectures. 
We advocate for designing an economical dataset distillation framework that is independent of the matching architectures.
With empirical observations, we argue that constraining the consistency of the real and synthetic image spaces will enhance the cross-architecture generalization. 
Motivated by this, we introduce Dataset Distillation via Disentangled Diffusion Model (D$^4$M), an efficient framework for dataset distillation. 
Compared to architecture-dependent methods, D$^4$M employs latent diffusion model to guarantee consistency and incorporates label information into category prototypes.
The distilled datasets are versatile, eliminating the need for repeated generation of distinct datasets for various architectures.
Through comprehensive experiments, D$^4$M demonstrates superior performance and robust generalization, surpassing the SOTA methods across most aspects.
\end{abstract}

%% file: sec/1_intro.tex
\section{Introduction}
\label{sec:intro}

The rapid growth in machine learning, resulting in large models and vast datasets, poses a challenge to researchers due to the escalating computational and storage demands.
\begin{figure}[!t]
  \centering
  \includegraphics[width=\linewidth]{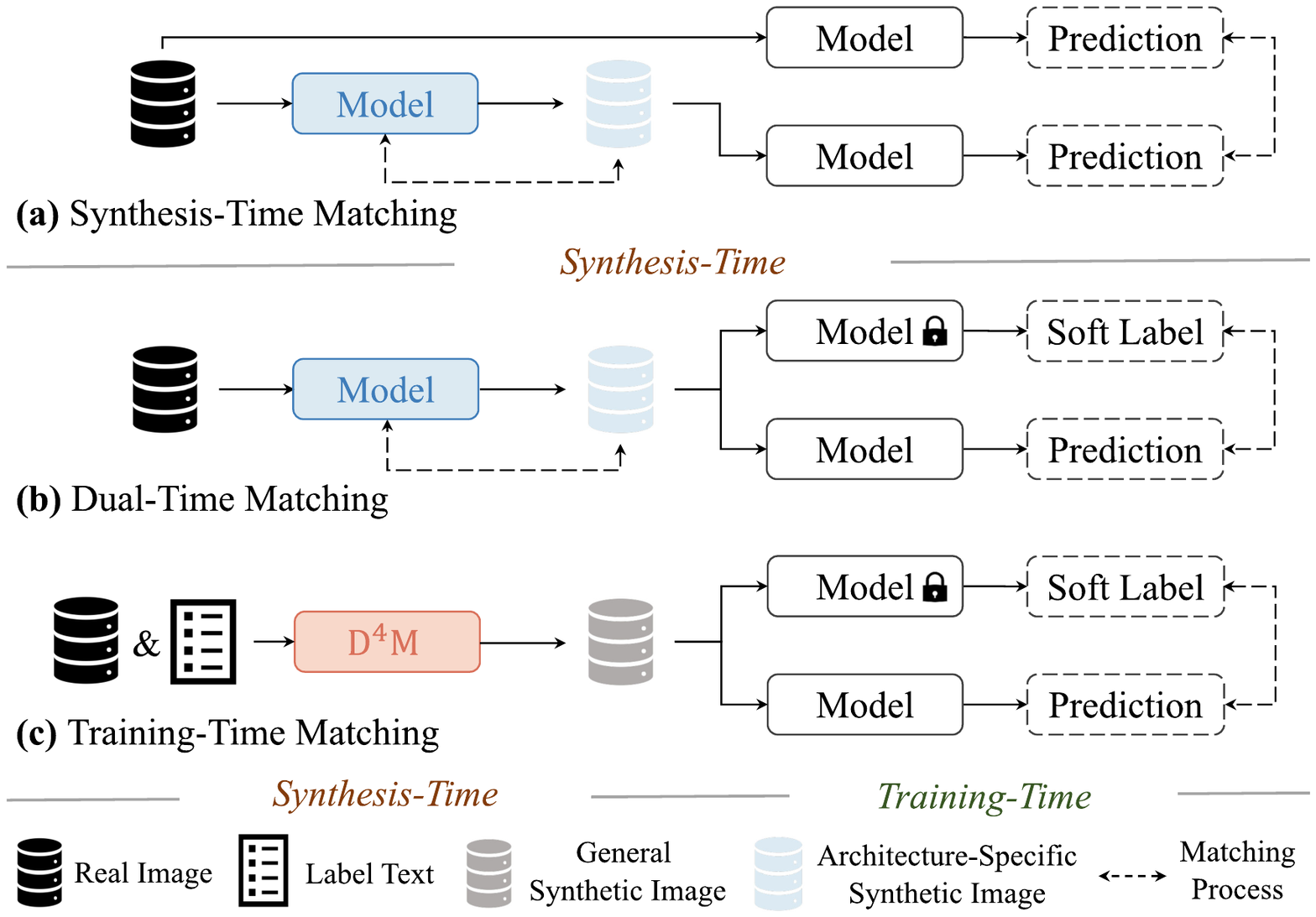}
  \caption{Comparison of various matching strategies in dataset distillation. (a) The bi-level optimization implements data matching at synthesis time. (b) Dual-Time Matching strategy decouples the bi-level optimization process into synthesis time and training time to save computational overhead. (c) D$^4$M utilizes multi-modal features (image and texts) to synthesize high-quality images. D$^4$M does not require matching process at Synthesis-Time.}
  \label{fig:main}
\end{figure}
Can the 'Divide-and-Conquer' algorithm~\cite{blahut2010fast} mitigate this challenge? 
From the perspective of dataset, recent research extends the coreset selection~\cite{chen2012super,toneva2018empirical,borsos2020coresets} to distillation techniques aimed at reducing dataset scales.
Dataset Distillation (DD) aims to synthesize a small dataset $\mathcal{S}$ from the original large-scale dataset $\mathcal{T}$, where $|\mathcal{S}| \ll |\mathcal{T}|$.
The information in $\mathcal{T}$ is condensed into a small dataset through DD.
Initially, the DD framework uses the bi-level optimization to generate datasets where the inner loop updates the network used for testing the classification performance and the outer loop synthesizes images according to matching strategies, such as gradient~\cite{zhao2020dataset,zhao2021dataset,lee2022dataset}, distribution~\cite{zhao2023dataset,wang2022cafe} or trajectory~\cite{cazenavette2022dataset,cui2023scaling}.

\begin{figure}[h]
    \centering
    \includegraphics[width=\linewidth]{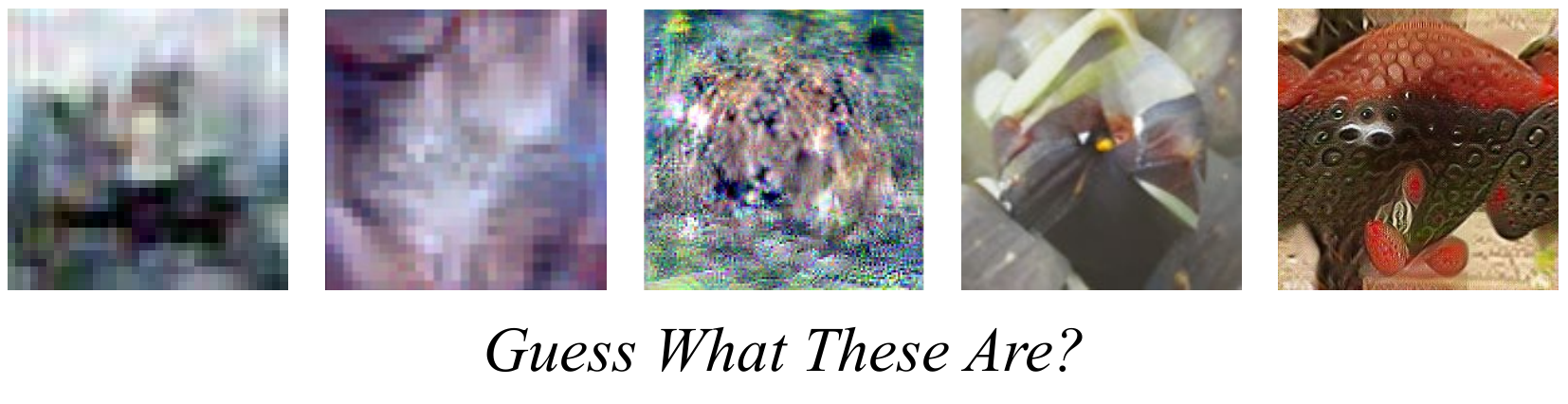}
    \caption{Visualizations of previous DD methods. Synthesis-Time Matching sacrifices part of the visual semantic expression in order to imitate the performance of the original dataset.}
    \label{fig:lack_semantic}
\end{figure}

Unfortunately, the existing solutions of DD mainly focus on small and simple datasets, such as CIFAR~\cite{krizhevsky2009learning} and MNIST~\cite{lecun2010mnist,xiao2017fashion}.
When it comes to large-scale and high-resolution datasets such as ImageNet~\cite{deng2009imagenet}, there exists unaffordable computational requirements and reduced performance.
Another challenge in DD is the cross-architecture generalization.
Previous methods conduct data matching within a fixed discriminative architecture, which makes the output space biased from the original image space.
As demonstrated in \cref{fig:lack_semantic}, this kind of dataset may be insightful for the networks but suffers from the lack of semantic information for humankind.
Furthermore, the dataset has to be distilled from scratch again and again to adapt to the emerging network architectures.
Obviously, these limitations constrain the scientific value and practical utility of the current solutions.
In this paper, we argue that an ideal DD method should meet the following properties.
% dsa_ship_cifar10；Tesla_cat_cifar10；MTT_lion_imgnt；glad_Fig_imgnt；srel_strawberry_imgnt

\textit{1) The synthesis process should not depend on a specific network architecture.}
Typically, a fixed architecture is required for data matching, which leads to low cross-architecture generalization performance because the output space is constrained by the architecture.
This problem arises once the matching process occurs in the synthesis time as shown in \cref{fig:main}(a) and (b).
Some work leverages a model pool instead of an individual matching model to alleviate this issue but makes the network hard to optimize~\cite{zhou2022dataset,wang2023dim}.
When the distillation process is architecture-free, there is no need to distill datasets for different architectures repeatedly.
In addition, constraining the consistency of input and output spaces will make the distilled images more realistic.
GlaD~\cite{cazenavette2023generalizing} seems to be a solution where the images are synthesized via Generative Adversarial Networks.
However, the synthetic images are still matched by the inner loop.

\textit{2) The method is capable of distilling datasets of various sizes and resolutions with limited computational resources.}
As illustrated in \cref{fig:main}(a), most DD solutions use bi-level optimization during synthesis time.
While the large-scale datasets are unable to perform a number of unrolled iterations on such a nested loop system.
Some works attempt to distill the ImageNet-1K but yield low testing accuracy~\cite{cazenavette2022dataset,cui2023scaling}.
A more effective method is depicted in \cref{fig:main}(b): the bi-level optimization is decoupled into synthesis time and training time~\cite{yin2023squeeze}.
However, the Dual-Time Matching (DTM) strategy leads to information loss at each stage, posing challenges for distillation on small datasets instead.

Inspired by these insights, we propose the \textbf{D}ataset \textbf{D}istillation via \textbf{D}isentangled \textbf{D}iffusion \textbf{M}odel (\textbf{D$^4$M}), an efficient approach designed for DD across varying sizes and resolutions as depicted in \cref{fig:main}(c).
In D$^4$M, the Synthesis-Time Matching (STM) is superseded by Training-Time Matching (TTM) which facilitates the fast distillation of large-scale datasets with constrained computational resources.
Furthermore, D$^4$M alleviates the architectural dependency and improves the cross-architecture generalization performance of the distilled dataset.
As the generative model, Diffusion Models ensure the consistency between input and output spaces, and its synthesis process does not rely on any specific matching architecture.
To mitigate the information loss due to insufficient data matching, the conditioning mechanism in Latent Diffusion Model (LDM) consistently infuses the semantic information of labels into the synthetic data during the denoising process.
The synthesis process of D$^4$M solely depends on the prototypes extracted from the original data, with synthesis speed scaling linearly with the size of datasets.
Moreover, the synthetic images exhibit realism at a high resolution of $512 \times 512$.
Our pivotal contributions are summarized as follows:
\begin{itemize}
    \item To the best of our knowledge, this is the first work that overcomes the pronounced dependency on specific architectures inherent in traditional DD frameworks.
    We introduce the TTM strategy, which paves the way for the generation of a curated and versatile distilled dataset.
    \item We propose D$^4$M that integrates the diffusion model into DD task for the first time.
    By leveraging label texts and the learned prototypes, we construct a multi-modal DD model that simultaneously enhances distillation efficiency and model performance.
    \item The method realizes the attainment of resolutions up to 512$\times$512 that exhibit high-fidelity and robust adaptability in the realm of DD. 
    This improvement is evidenced across a spectrum of datasets, extending from the ImageNet-1K to CIFAR-10/100.
    \item We conduct extensive experiments and ablation studies.
    The results outperform the SOTA in most cases, substantiating the superior performance, computational efficiency, and robustness of our method.
\end{itemize}

%% file: sec/2_related.tex
\section{Related Work}
\label{sec:rw}
\begin{figure*}[t]
  \centering
  \includegraphics[width=0.9\linewidth]{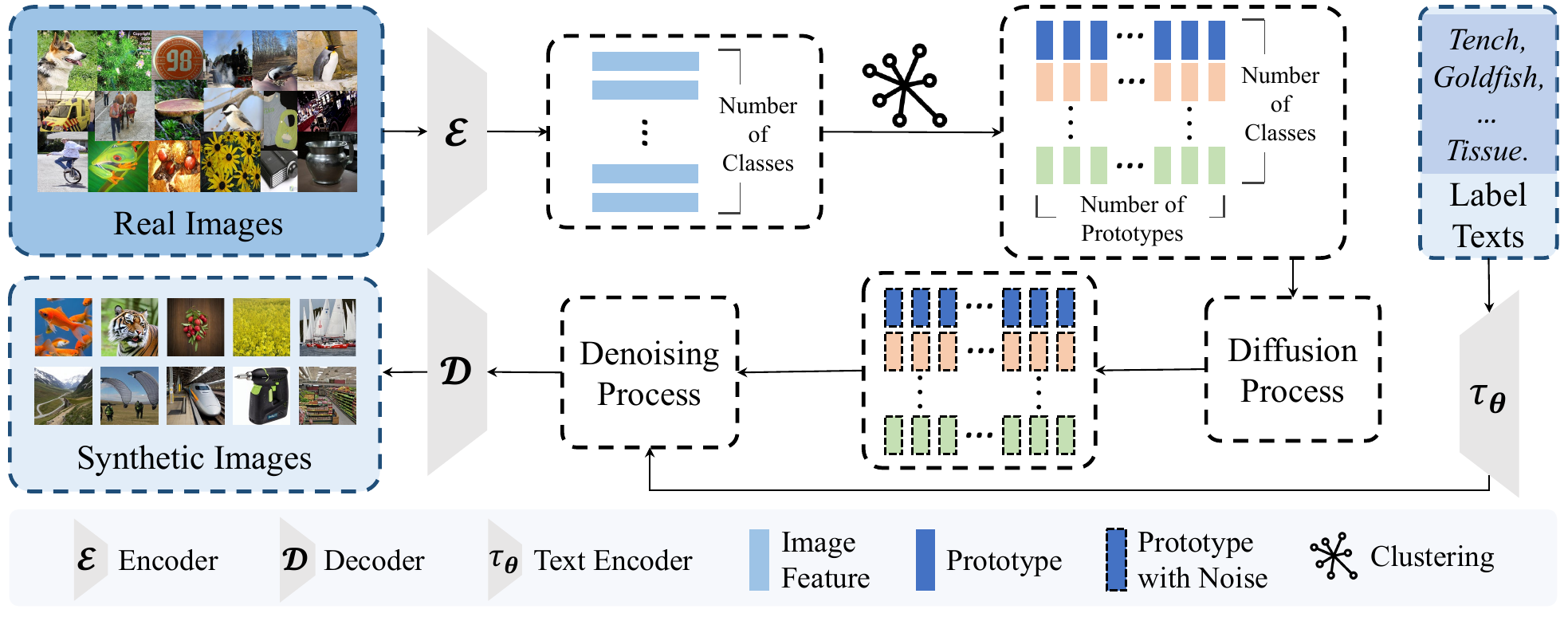}
  \caption{\textbf{Pipeline of Dataset Distillation via Disentangled Diffusion Model (D$^4$M)}. Rather than using the embedded features directly, D$^4$M disentangles feature extraction from image generation in diffusion models through prototype learning.}
  \label{fig:d4m}
\end{figure*}

\subsection{Dataset Distillation}
The existing DD approaches are taxonomized into meta-learning matching and data matching frameworks~\cite{yu2023dataset,sachdeva2023data,geng2023survey,lei2023comprehensive}.
The meta-learning matching aims to optimize the meta-test loss on real dataset for the model meta-trained by the distilled dataset.
The gradients are back-propagated to supervise the DD directly~\cite{wang2018dataset,nguyen2020dataset,nguyen2021dataset,deng2022remember,loo2022efficient,zhou2022dataset}.

Unlike optimizing the performance on the DD explicitly, data matching encourages the consistency between the same network architecture trained by distilled and real dataset.
Matching the gradients generated by the networks is a reliable surrogate task~\cite{zhao2020dataset,zhao2021dataset,lee2022dataset,jiang2023delving}. 
Matching Training Trajectory (MTT)~\cite{cazenavette2022dataset,du2023minimizing} is then proposed to solve the issue that errors are accumulated during validation in gradient matching.
TESLA~\cite{cui2023scaling} reduced the complexity of gradients calculating with constant memory, allowing DD to be achieved in ImageNet for the first time.
% Besides, distribution matching optimizes the distance between the two distributions using some metrics directly, such as MMD~\cite{zhao2023dataset} and CAFE~\cite{wang2022cafe}.
Besides, distribution matching optimizes the distance between the two distributions, such as MMD~\cite{zhao2023dataset} and CAFE~\cite{wang2022cafe}.

The aforementioned methods only implement various matching strategies at synthesis time.
SRe$^2$L~\cite{yin2023squeeze} argues that decoupling the bi-level optimization into Squeeze, Recover, and Relabel leads to a good performance on large-scale datasets.
Inspired by this, we summarize previous works into STM and DTM.
D$^4$M implements the TTM with the help of soft labels, which is considered a feature distribution matching approach.

\subsection{Diffusion Models}
% Given observed samples $x$ drawn from a target distribution, generative models aim to approximate the real distribution $P(x)$ and thus the model can produce novel samples from the approximated distribution.
The Diffusion Model has demonstrated remarkable capabilities within the generative models.
Given samples $x$ observed from a target distribution, the goal of generative models is approximating the true distribution $P(x)$, enabling the generation of novel samples from it.
Denoising Diffusion Probabilistic Models (DDPM)~\cite{ho2020denoising} aims to learn a reverse process of a fixed Markov Chain for generating images.
However, DDPM is expensive to optimize and evaluate in the original pixel space.

Latent Diffusion Model (LDM)~\cite{rombach2022high}, a recent state-of-the-art diffusion model, addresses this by abstracting high-frequency, imperceptible details into a compact latent space, thereby streamlining both training and inference.
LDM has been applied in image editing~\cite{takagi2023high, wu2023latent}, video processing~\cite{blattmann2023align, esser2023structure}, audio generation~\cite{shen2023difftalk, jeong2023power} and 3D model reconstruction~\cite{kim2023neuralfield, koo2023salad, lyu2023controllable, chen2023executing}.
Notably, the proficiency of LDM in abstracting and generating images within the latent space exactly resonates with the foundational tenets of DD.

%% file: sec/3_method.tex
\section{Method}
% This section reviews the objective of the diffusion model and elucidates the rationale behind its decoupling. 
% Furthermore, we introduce the TTM strategy.

\subsection{Preliminaries on Diffusion Models}
A pivotal step in DD is the generation of the distilled images.
Distinct from the data-matching approaches, our method harnesses the prior knowledge embedded in the pre-trained generative models, offering a high-quality initialization for TTM.
Recently, diffusion models have emerged as SOTA in generative models~\cite{yang2022diffusion,luo2022understanding}.
As aforementioned, the synthesis process of the diffusion model does not rely on any specific matching architecture, ensuring the consistency between input and output spaces.
For a sequence of denoising autoencoders $\epsilon_\theta$, the training objective of Denoising Diffusion Probability Model (DDPM)~\cite{ho2020denoising} is defined as
\begin{equation}
L_{DM}=\mathbb{E}_{x, \epsilon \sim \mathcal{N}(0,1), t}\left[\left\|\epsilon-\epsilon_\theta\left(x_t, t\right)\right\|_2^2\right],
\label{eq:dm}
\end{equation}
with the timestamp $t$ uniformly sampled from $\{1,\dots,T\}$.
Although the DDPM does not cater to our goal of synthesizing images within the condensed features, we turn our attention to the LDM~\cite{rombach2022high}. 

LDM effectively compresses the working space from the original pixel space $x$ to a more compact latent space $z$. 
Such a transition is close to our intent of encapsulating images into condensed features.
LDM constructs an optimized low-dimensional latent space by training a perceptual compression model composed of the encoder ($\mathcal{E}$) and decoder ($\mathcal{D}$). 
This latent space effectively abstracts high-frequency imperceptible details than pixel space~\cite{rombach2022high}.
In this case, the objective function with text encoder $\tau_\theta$ is redefined as
\begin{equation}
L_{LDM}=\mathbb{E}_{\mathcal{E}(x), y, \epsilon \sim \mathcal{N}(0,1), t}\left[\left\|\epsilon-\epsilon_\theta\left(z_t, t, \tau_\theta(y)\right)\right\|_2^2\right].
\label{eq:ldm}
\end{equation}

\subsection{Disentangled Diffusion Model}
The existing diffusion methods are capable of generating high-quality images directly from the given images and prompts. 
However, it is imperative for the DD model to aggregate the given images into a few condensed features before synthesis. 
The images in the original dataset encapsulate a spectrum of information from low-level texture patterns to high-level semantic information, along with potential redundancies. 
Since the diffusion models do not have the capability of aggregating this information among images, it is necessary to extract the salient feature representative of each category before employing the generative model.
Consequently, it is essential to disentangle the diffusion models.

Employing prototypes in standard classification tasks offers the benefit of addressing the open-world recognition challenge, thereby enhancing the robustness of models~\cite{yang2018robust, li2021adaptive,zhang2021prototype}. 
Therefore, initializing the input of the diffusion model with prototypes not only reduces data redundancy but also elevates the quality of the distilled dataset.
As illustrated in \cref{fig:d4m}, we leverage the pre-trained autoencoder $\mathcal{E}$ inherent in the LDM to extract feature representations from original images.
Subsequently, we perform a clustering algorithm to calculate the cluster centers as prototypes for each category.
Given the considerable size of the original dataset, we adopt the Mini-Batch $k$-Means~\cite{sculley2010web} to mitigate the memory overhead of large-scale clustering.
This approach iteratively optimizes a mini-batch of samples in each step, accelerating the clustering process with a minimal compromise in accuracy.

% We optimize 10/50 prototypes for each category.
Specifically, the clustering algorithm consists of two primary steps: assignment $z$
\begin{align}
\label{eq:assign}
    & z^c \leftarrow z \\
    \text{s.t.}~&\mathop{\arg\min}\limits_{c}\|z-z^c\|^2, c=1,\dots,C
\end{align}
and update $z^c$
\begin{equation}
\label{eq:update}
    z^c \leftarrow (1-\eta)z^c+\eta z.
\end{equation}
Here $z$ is the latent variable generated by $\mathcal{E}$, and $z^c$ represents the cluster centers (prototypes), $C$ is the number of cluster centers.
The learning rate $\eta$ is often calculated by $\frac{1}{|z^c|}$.
Ultimately, we employ the prototypes $\Bar{Z}=\{z^c_l | c=1,\dots,C,~~ l=1,\dots,L\}$ from all categories as input to the diffusion process for image synthesis.

Moreover, LDM is capable of modeling the conditional distribution, enabling DD tasks to incorporate the label information into synthetic images.
In \cref{eq:ldm}, LDM introduces a domain-specific encoder $\tau_\theta$ to map the textual labels (prompts) into the feature space.
This mapping is seamlessly integrated into the U-Net architecture ($\mathcal{U}_t$) through a cross-attention layer, facilitating the fusion of multi-modal features.
For each prototype $z^c$ and its corresponding label $L$, the synthesis process is formulated as 
\begin{equation}
    output=\mathcal{D}(\mathcal{U}_t(Concat(z^c_t,\tau_\theta(L)))
\end{equation}
where $z_t^c$ represents the $c$-th prototype with noise.
The distillation process is summarized in \cref{alg:d4m}.
\begin{algorithm}[!t]
\caption{\textbf{D}ataset \textbf{D}istillation via \textbf{D}isentangled \textbf{D}iffusion \textbf{M}odel (\textbf{D$^4$M})}
\begin{algorithmic}[1]
    \REQUIRE ($\mathcal{T}$,$\mathcal{L}$): Real images and their label texts. \\
    % \REQUIRE $\mathcal{L}$: Label texts. \\
    \REQUIRE $\mathcal{E}$: Pre-trained encoder. \\
    \REQUIRE $\mathcal{D}$: Pre-trained decoder. \\
    \REQUIRE $\tau_\theta$: Pre-trained text encoder. \\
    \REQUIRE $\mathcal{U}_t$: Pre-trained time-conditional U-Net. 
    \REQUIRE C: Number of prototypes.
    % \REQUIRE N: Number of classes.
    \STATE $Z=\mathcal{E}(\mathcal{T})\sim P_z$
    \hfill $\rhd$ Compressed latent space \\
    \STATE \textbf{for each} $L\in\mathcal{L}$ \textbf{do}
    \STATE \enspace \quad \textbf{for} mini-batch $z\in L$ \textbf{do} 
    \STATE \qquad \enspace \enspace $z^c\sim P_z, c=1,\dots,C$ \\
    \hfill $\rhd$ Initialize cluster centers
    \STATE \qquad \enspace \enspace $z^c \leftarrow z, \text{s.t.}\mathop{\arg\min}\limits_{c}\|z-z^c\|^2$
    \hfill $\rhd$ Assignment \\
    \STATE \qquad \enspace \enspace $\eta=\frac{1}{|z^c|}$ 
    \hfill $\rhd$ Update learning rate
    \STATE \qquad \enspace \enspace $z^c \leftarrow (1-\eta)z^c+\eta z$
    \hfill $\rhd$ Update \\
    \STATE \enspace \quad \textbf{end for}
    \STATE \enspace \quad $y=\tau_\theta(L)$
    \hfill $\rhd$ Label text embedding \\
    \STATE \enspace \quad \textbf{for each} $z^c$ \textbf{do}
    \STATE \qquad \enspace \enspace $z_t^c\sim q(z_t^c|z^c)$
    \hfill $\rhd$ Diffusion process \\
    \STATE \qquad \enspace \enspace $\tilde{z}^c=\mathcal{U}_t(Concat(z_t^c,y))$
    \hfill $\rhd$ Denoising process \\
    \STATE \enspace \quad \textbf{end for}
    \STATE \textbf{end for}
    \STATE $\mathcal{S}=\mathcal{D}(\tilde{Z}^c)$
    \hfill $\rhd$ Generate image
    \ENSURE $\mathcal{S}$: Distilled images.
\end{algorithmic}
\label{alg:d4m}
\end{algorithm}

\begin{figure*}
  \centering
  \begin{tabular}{*{6}{@{\hspace{0.003\linewidth}}c}@{\hspace{0.012\linewidth}}c*{5}{@{\hspace{0.003\linewidth}}c}}
    \multirow{2}{*}{\rotatebox{90}{\scriptsize{ImageNet-1K}}}
    & \includegraphics[width=0.09\linewidth]{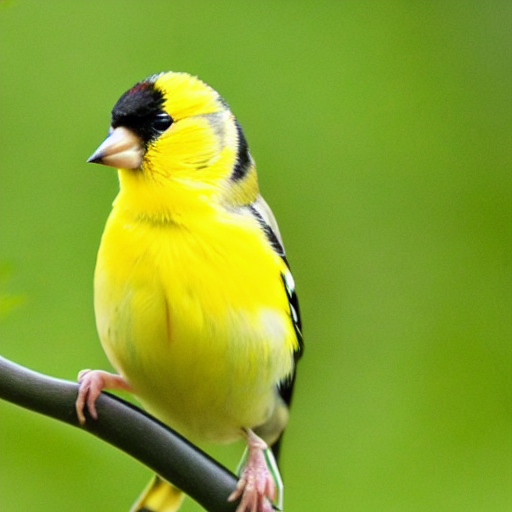}
    & \includegraphics[width=0.09\linewidth]{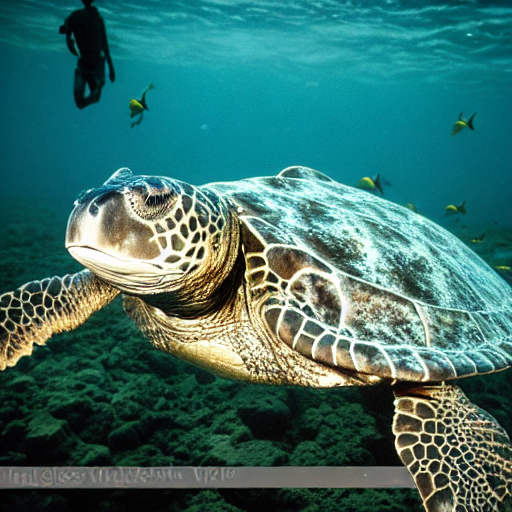} 
    & \includegraphics[width=0.09\linewidth]{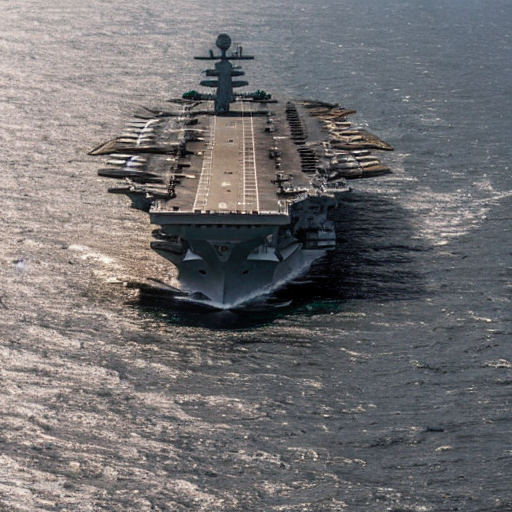} 
    & \includegraphics[width=0.09\linewidth]{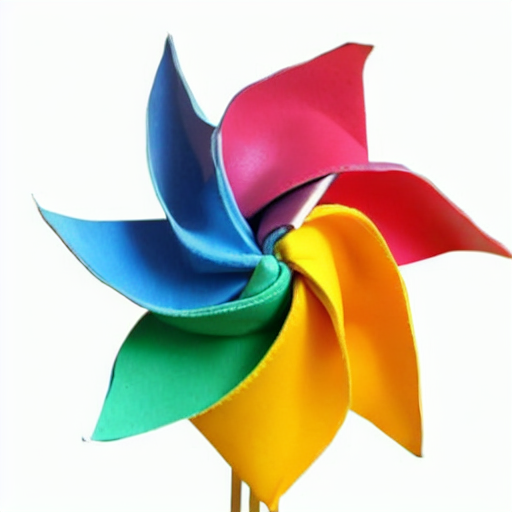} 
    & \includegraphics[width=0.09\linewidth]{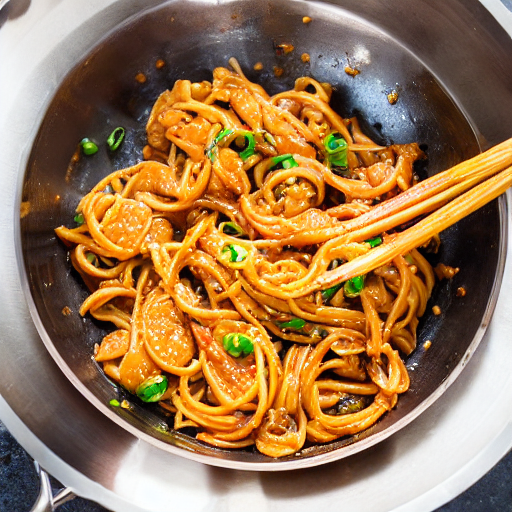}
    & \multirow{2}{*}{\rotatebox{90}{\scriptsize{Tiny-ImageNet}}}
    & \includegraphics[width=0.09\linewidth]{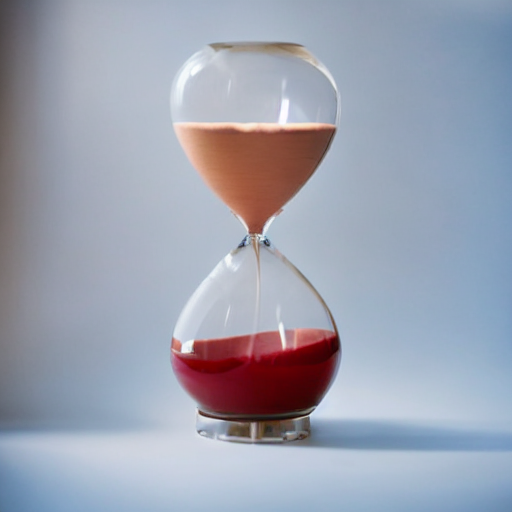} 
    & \includegraphics[width=0.09\linewidth]{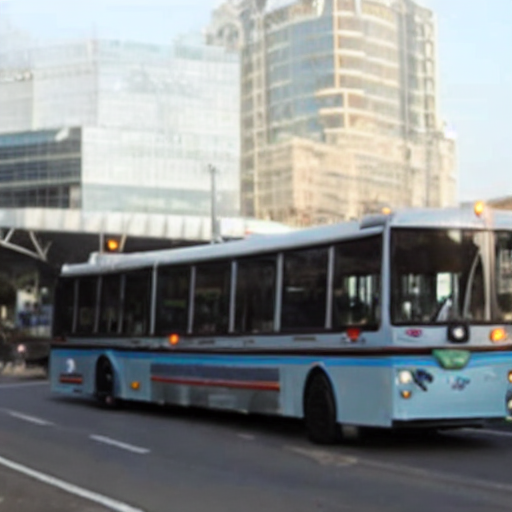} 
    & \includegraphics[width=0.09\linewidth]{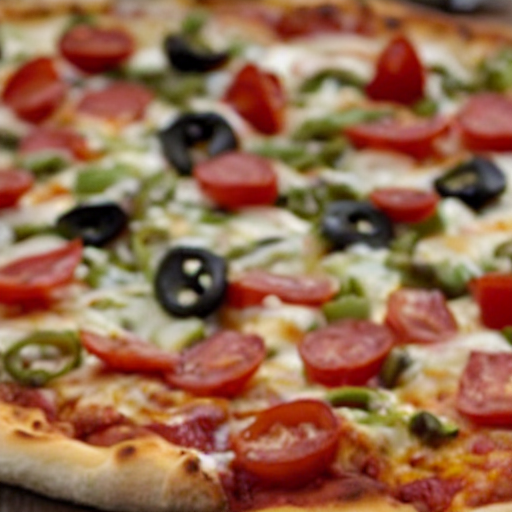} 
    & \includegraphics[width=0.09\linewidth]{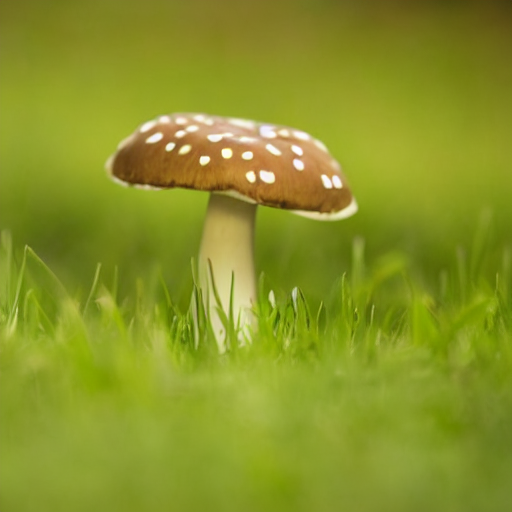} 
    & \includegraphics[width=0.09\linewidth]{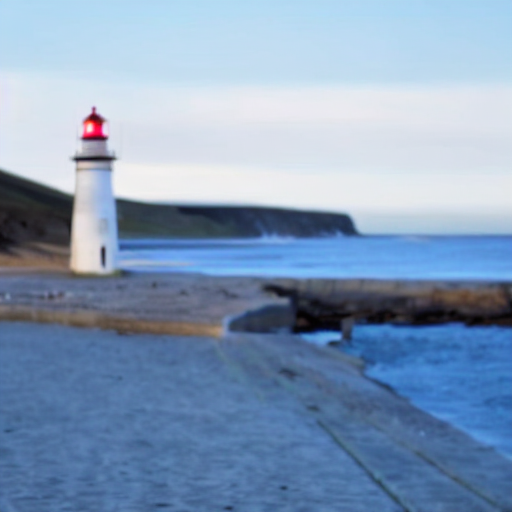}
    \\
    & \includegraphics[width=0.09\linewidth]{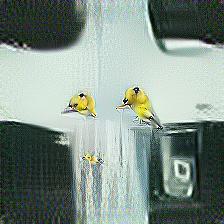} 
    & \includegraphics[width=0.09\linewidth]{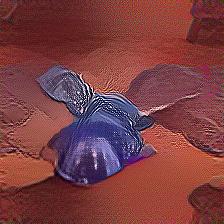} 
    & \includegraphics[width=0.09\linewidth]{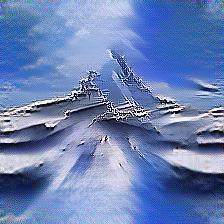} 
    & \includegraphics[width=0.09\linewidth]{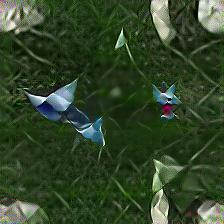} 
    & \includegraphics[width=0.09\linewidth]{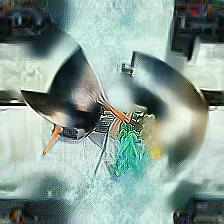}
    & 
    & \includegraphics[width=0.09\linewidth]{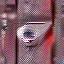}
    & \includegraphics[width=0.09\linewidth]{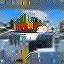} 
    & \includegraphics[width=0.09\linewidth]{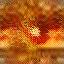} 
    & \includegraphics[width=0.09\linewidth]{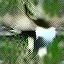} 
    & \includegraphics[width=0.09\linewidth]{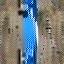} \\
    & Goldfinch & Turtle & Carrier & Pinwheel & Wok & & Hourglass & Trolleybus & Pizza & Mushroom & Beacon \\
    &&&&&&&&& \\
    \multirow{2}{*}{\rotatebox{90}{\scriptsize{CIFAR-10}}}
    & \includegraphics[width=0.09\linewidth]{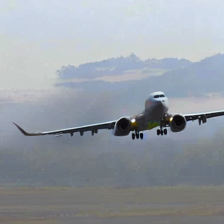}
    & \includegraphics[width=0.09\linewidth]{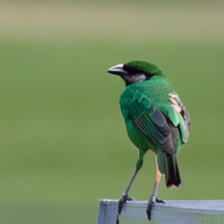} 
    & \includegraphics[width=0.09\linewidth]{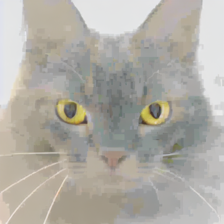} 
    & \includegraphics[width=0.09\linewidth]{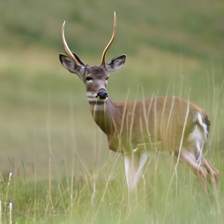} 
    & \includegraphics[width=0.09\linewidth]{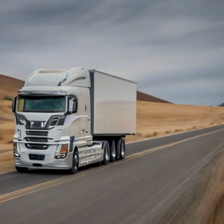} 
    & \multirow{1}{*}{\rotatebox{90}{\scriptsize{CIFAR-100}}}
    & {\includegraphics[width=0.09\linewidth]{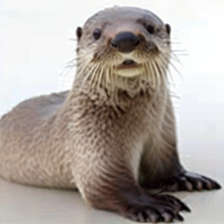}}
    & \includegraphics[width=0.09\linewidth]{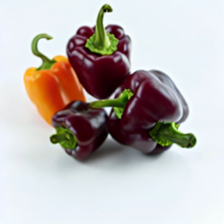} 
    & \includegraphics[width=0.09\linewidth]{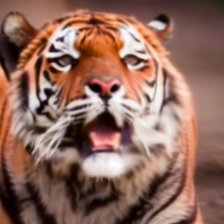} 
    & \includegraphics[width=0.09\linewidth]{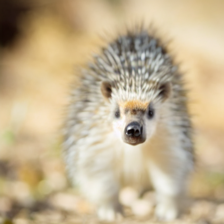} 
    & \includegraphics[width=0.09\linewidth]{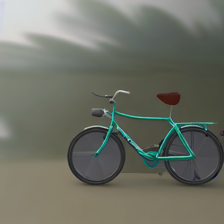} \\
    & \includegraphics[width=0.09\linewidth]{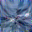} 
    & \includegraphics[width=0.09\linewidth]{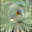} 
    & \includegraphics[width=0.09\linewidth]{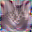} 
    & \includegraphics[width=0.09\linewidth]{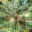} 
    & \includegraphics[width=0.09\linewidth]{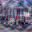}
    & 
    & \includegraphics[width=0.09\linewidth]{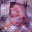}
    & \includegraphics[width=0.09\linewidth]{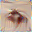} 
    & \includegraphics[width=0.09\linewidth]{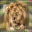} 
    & \includegraphics[width=0.09\linewidth]{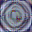} 
    & \includegraphics[width=0.09\linewidth]{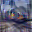} \\
    & Airplane & Bird & Cat & Deer & Truck & & Otter & Peppers & Tiger & Porcupine & Bicycle \\
  \end{tabular}
  \caption{Visualization results. The top row of each dataset comes from D$^4$M and the bottom comes from SRe$^2$L~\cite{yin2023squeeze} (ImageNet-1K and Tiny-ImageNet) and MTT~\cite{cazenavette2022dataset} (CIFAR-10/100). The images generated by D$^4$M have better resolution and are more lifelike.}
  \label{fig:vis_1}
\end{figure*}

\begin{figure}
  \centering
  \begin{tabular}{*{8}{@{\hspace{0.003\linewidth}}c}}
    \includegraphics[width=0.118\linewidth]{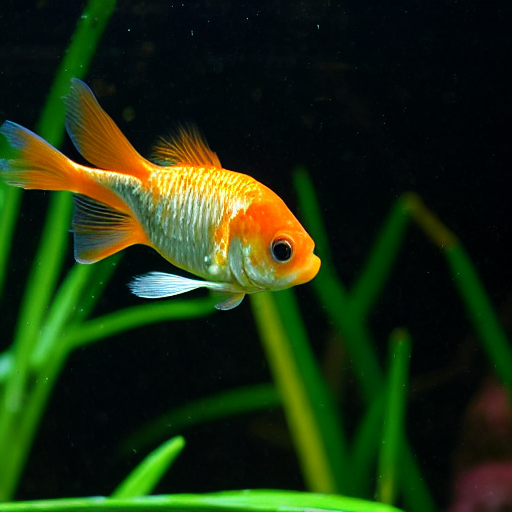}
    & \includegraphics[width=0.118\linewidth]{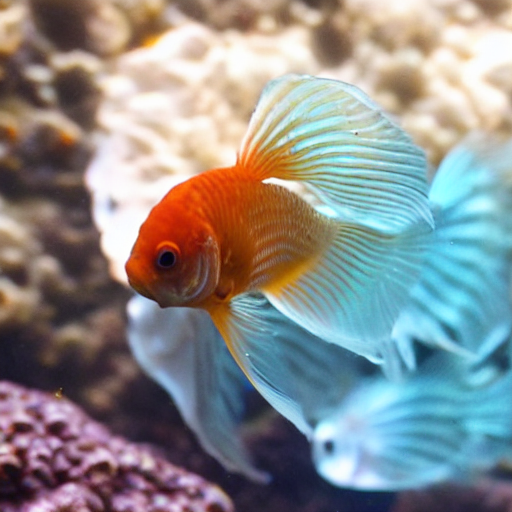} 
    & \includegraphics[width=0.118\linewidth]{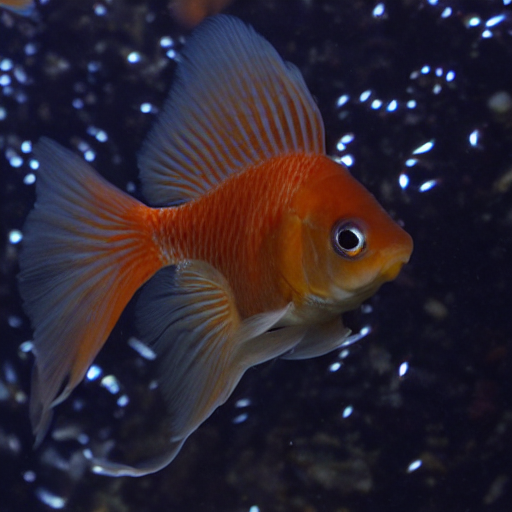} 
    & \includegraphics[width=0.118\linewidth]{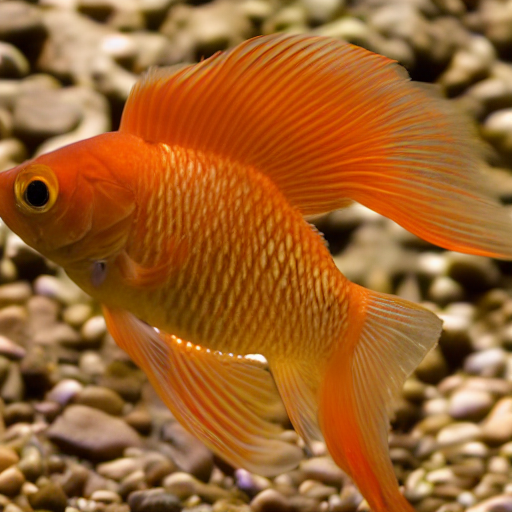} 
    & \includegraphics[width=0.118\linewidth]{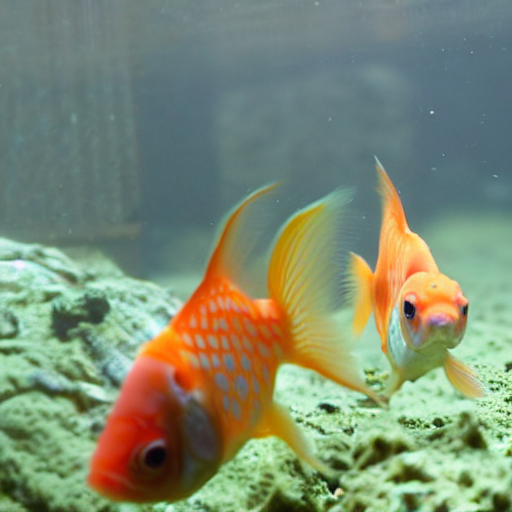} 
    & \includegraphics[width=0.118\linewidth]{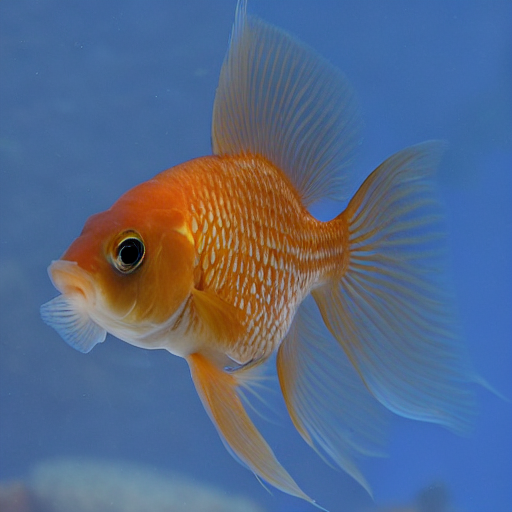} 
    & \includegraphics[width=0.118\linewidth]{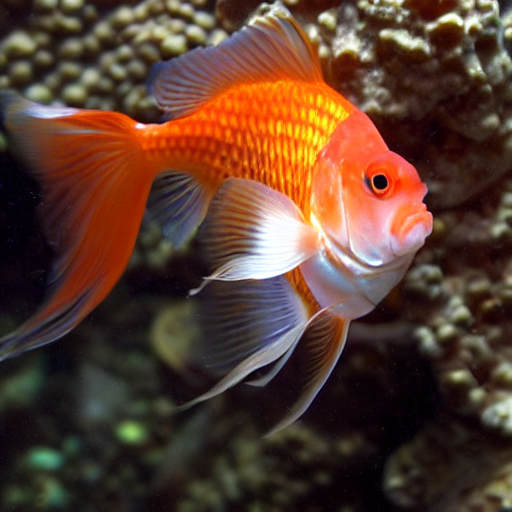} 
    & \includegraphics[width=0.118\linewidth]{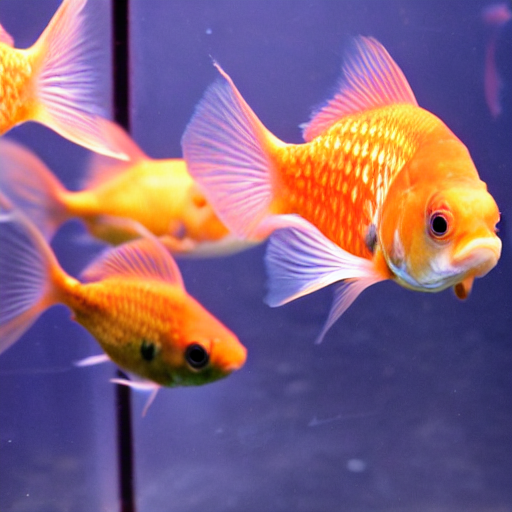} \\
    % & \includegraphics[width=0.09\linewidth]{imgs/vis_fish/1-image8.png} 
    % & \includegraphics[width=0.09\linewidth]{imgs/vis_fish/1-image9.png} \\
    \includegraphics[width=0.118\linewidth]{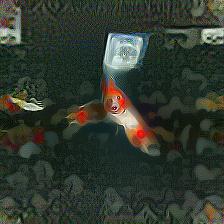} 
    & \includegraphics[width=0.118\linewidth]{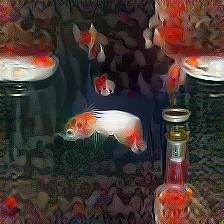} 
    & \includegraphics[width=0.118\linewidth]{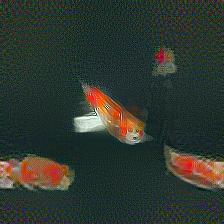} 
    & \includegraphics[width=0.118\linewidth]{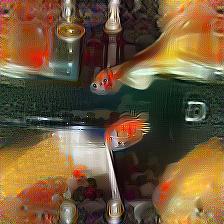} 
    & \includegraphics[width=0.118\linewidth]{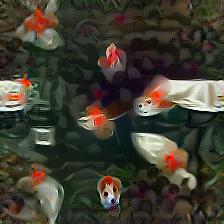} 
    & \includegraphics[width=0.118\linewidth]{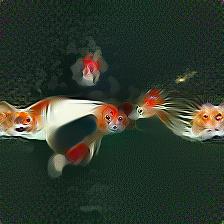} 
    & \includegraphics[width=0.118\linewidth]{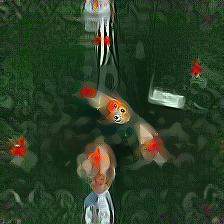} 
    & \includegraphics[width=0.118\linewidth]{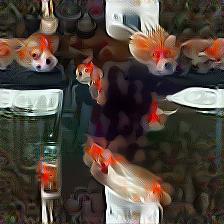} \\
    % & \includegraphics[width=0.09\linewidth]{imgs/vis_fish/class001_id008.jpg} 
    % & \includegraphics[width=0.09\linewidth]{imgs/vis_fish/class001_id009.jpg} \\
  \end{tabular}
  \caption{Visualization results within one category. D$^4$M (top) provides richer semantic information than SRe$^2$L.}
  \label{fig:vis_2}
\end{figure}

\subsection{Training-Time Matching}
Since eliminates the necessity of matching with a specific architecture, separating data matching from the synthesis process reduces the computational overhead on large-scale datasets and addresses the cross-architecture issue inherent in the STM strategy.
However, based on previous research~\cite{yin2023squeeze,cazenavette2022dataset,cui2023scaling} and preliminary experiments, we find that training large-scale distilled datasets with hard labels is prone to low testing accuracy.
To address this, we introduce the TTM strategy, which is considered a distribution matching approach.

TTM refers to training on distilled datasets with soft labels.
Label softening is widely adapted in distillation tasks~\cite{hinton2015distilling,muller2019does,yim2017gift}.
Since D$^4$M infuses the label features into the synthetic data, it is natural to use the soft label during TTM.
We employ soft label to align the distribution of student prediction $S_\theta(x)$ with teacher network $T$:
\begin{equation}
\theta_\text{student}=\mathop{\arg\min}\limits_{\theta\in\Theta}L_{KL}(T(x),S_{\theta}(x))
\end{equation}
where $T(x)$/$S_\theta(x)$ is the teacher/student prediction for the distilled image $x$ and $L_{KL}$ represents the KL divergence.
The output of the teacher network, also known as soft prediction or soft label, encapsulates richer semantic information compared to hard labels.
Matching with the soft labels during training will enhance the robustness and generalization capability of the trained model~\cite{hinton2015distilling}.
For a fair comparison, we use the soft label storage method similar to the FKD~\cite{shen2022fast} method, which generates soft labels and conducts matching at each training epoch:
\begin{equation}
\theta_\text{student}^{t+1}=\mathop{\arg\min}\limits_{\theta\in\Theta}L_{KL}(T^t(x),S_{\theta}^t(x)).
\end{equation}
% \Cref{sec:ts-anal} provides a detailed analysis of the soft labels generated by different teacher networks.

\begin{table*}
\begin{tabular}{lccccccccc}
\toprule
\multirow{2}{*}{Dataset} & \multirow{2}{*}{IPC} & \multicolumn{2}{c}{Meta-Learning} & \multicolumn{3}{c}{Data-Matching} & & & \multirow{2}{*}{Full Dataset} \\
 &  & KIP & FRePO & DSA & CAFE & TESLA & SRe$^2$L$^\dagger$ & D$^4$M &  \\
\midrule
\multirow{2}{*}{CIFAR-10} & 10 & 62.7$\pm$0.3 & 65.5$\pm$0.6 & 52.1$\pm$0.5 & 50.9$\pm$0.5 & \textbf{66.4$\pm$0.8} & \multirow{2}{*}{(60.2)} & \cellcolor{lightgray!30}56.2$\pm$0.4 & \cellcolor{lightgray!30} \\
 & 50 & 68.6$\pm$0.2 & 71.7$\pm$0.2 & 60.6$\pm$0.5 & 62.3$\pm$0.4 & 72.6$\pm$0.7 &  & \cellcolor{lightgray!30}\textbf{72.8$\pm$0.5} & \cellcolor{lightgray!30}\multirow{-2}{*}{84.8$\pm$0.1} \\
\midrule
\multirow{2}{*}{CIFAR-100} & 10 & 28.3$\pm$0.1 & 42.5$\pm$0.2 & 32.3$\pm$0.3 & 31.5$\pm$0.2 & 41.7$\pm$0.3 & - & \cellcolor{lightgray!30}\textbf{45.0$\pm$0.1} & \cellcolor{lightgray!30} \\
 & 50 & - & 44.3$\pm$0.2 & 42.8$\pm$0.4 & 42.9$\pm$0.2 & 47.9$\pm$0.3 & - & \cellcolor{lightgray!30}\textbf{48.8$\pm$0.3} & \cellcolor{lightgray!30}\multirow{-2}{*}{56.2$\pm$0.3} \\
\bottomrule
\end{tabular}
\caption{\textbf{Top-1 Accuracy$\uparrow$ on small datasets}. We train the ConvNet-W128~\cite{gidaris2018dynamic} from scratch 5 times on the distilled dataset and evaluate them on the original test dataset to get the $\Bar{x}\pm std$. $\dagger$: SRe$^2$L~\cite{yin2023squeeze} achieves 60.2\% Top-1 Accuracy on CIFAR-10 with IPC-1K.}
\label{tab:small_data}
\end{table*}

%% file: sec/4_experiment.tex
\section{Experiments}
\subsection{Setting and Evaluation}
% \textcolor{green}{
We evaluate the performance of D$^4$M across various datasets and networks. 
All models employed for ImageNet-1K and Tiny-ImageNet are sourced from the PyTorch official model repository, while the ConvNet utilized for CIFAR-10/100 is based on the architecture proposed by Gidaris \etal~\cite{gidaris2018dynamic}. 
Performance validation was carried out using PyTorch on NVIDIA V100 GPUs. Detailed training and validation hyperparameters are available in the supplementary material.
% }

\subsection{Dataset Distillation Results}
In our comparative analysis, we evaluate the D$^4$M against a range of techniques, encompassing both meta-learning and data-matching strategies. 
For small datasets, our comparison included two meta-learning methods: KIP~\cite{nguyen2021dataset} and FRePO~\cite{zhou2022dataset}, alongside four data-matching techniques: DSA~\cite{zhao2021dataset}, CAFE~\cite{wang2022cafe}, TESLA~\cite{cui2023scaling}, and SRe$^2$L~\cite{yin2023squeeze}. 
In the context of large-scale datasets, our focus shifted to a detailed comparison between TESLA and SRe$^2$L.

{\bf CIFAR-10 and CIFAR-100}
For small dataset distillation, the STM strategy outperforms when the number of categories and IPC (Image Per Class) are limited. 
However, as the category increases, the TTM strategy becomes more effective. 
This shift is attributed to the fact that the optimal solution derived from STM fails to ensure the convergence of the network training with large category numbers, thereby capping the testing performance. 
As evidenced in \cref{tab:small_data}, when applied to CIFAR-100, D$^4$M attains a Top-1 accuracy of 45.0\% with merely IPC-10. 
This performance surpasses that of FRepo and TESLA by 2.5\% and 3.3\%.

\begin{table}
  \centering
  \begin{tabular}{lccccc}
    \toprule
    Dataset & IPC & Method & R18 & R50 & R101 \\
    \midrule
    \multirow{10}{*}{ImageNet-1K} & \multicolumn{2}{c}{Full Dataset$^\dagger$} & \cellcolor{lightgray!30}69.8 & \cellcolor{lightgray!30}80.9 & \cellcolor{lightgray!30}81.9 \\
    & \multirow{3}{*}{10} & TESLA & 7.7 & - & - \\
       &  & SRe$^2$L & 21.3 & 28.4 & 30.9 \\
       &  & D$^4$M & \cellcolor{lightgray!30}\textbf{27.9} & \cellcolor{lightgray!30}\textbf{33.5} & \cellcolor{lightgray!30}\textbf{34.2} \\
       & \multirow{2}{*}{50} & SRe$^2$L & 46.8 & 55.6 & 60.8 \\
       &  & D$^4$M & \cellcolor{lightgray!30}\textbf{55.2} & \cellcolor{lightgray!30}\textbf{62.4} & \cellcolor{lightgray!30}\textbf{63.4} \\
       & \multirow{2}{*}{100} & SRe$^2$L & 52.8 & 61.0 & 65.8 \\
       &  & D$^4$M & \cellcolor{lightgray!30}\textbf{59.3} & \cellcolor{lightgray!30}\textbf{65.4} & \cellcolor{lightgray!30}\textbf{66.5} \\
       & \multirow{2}{*}{200} & SRe$^2$L & 57.0 & 64.6 & 65.9 \\
       &  & D$^4$M & \cellcolor{lightgray!30}\textbf{62.6} & \cellcolor{lightgray!30}\textbf{67.8} & \cellcolor{lightgray!30}\textbf{68.1} \\
    \midrule
       \multirow{7}{*}{Tiny-ImageNet} & \multicolumn{2}{c}{Full Dataset$^\ddagger$} & \cellcolor{lightgray!30}61.9 & \cellcolor{lightgray!30}62.0 & \cellcolor{lightgray!30}62.3 \\
       & \multirow{3}{*}{50} & SRe$^2$L & 44.0 & 47.7 & 49.1 \\
       &  & D$^4$M & \cellcolor{lightgray!30}46.2 & \cellcolor{lightgray!30}51.8 & \cellcolor{lightgray!30}51.0 \\
       &  & D$^4$M-G & \cellcolor{lightgray!30}\textbf{46.8} & \cellcolor{lightgray!30}\textbf{51.9} & \cellcolor{lightgray!30}\textbf{53.2} \\
       & \multirow{3}{*}{100} & SRe$^2$L & 50.8 & 53.5 & 54.2 \\
       &  & D$^4$M & \cellcolor{lightgray!30}51.4 & \cellcolor{lightgray!30}54.8 & \cellcolor{lightgray!30}\textbf{55.3} \\
       &  & D$^4$M-G & \cellcolor{lightgray!30}\textbf{53.3} & \cellcolor{lightgray!30}\textbf{54.9} & \cellcolor{lightgray!30}54.5 \\
    \bottomrule
  \end{tabular}
  \caption{\textbf{Top-1 Accuracy$\uparrow$ on large-scale datasets}. SRe$^2$L~\cite{yin2023squeeze} and our D$^4$M employ ResNet18 as the teacher model to generate the soft label while TESLA~\cite{cui2023scaling} uses the ConvNetD4. All standard deviations in this table are $<1$. $\dagger$: The results of ImageNet-1K come from the official PyTorch \href{https://pytorch.org/vision/main/models/resnet.html}{websites}. $\ddagger$: The results of Tiny-ImageNet come from the model trained from scratch with the official PyTorch \href{https://github.com/pytorch/vision/tree/main/references/classification}{code}.}
  \label{tab:large_data}
\end{table}

\begin{figure*}[t]
  \centering
  \includegraphics[width=\linewidth]{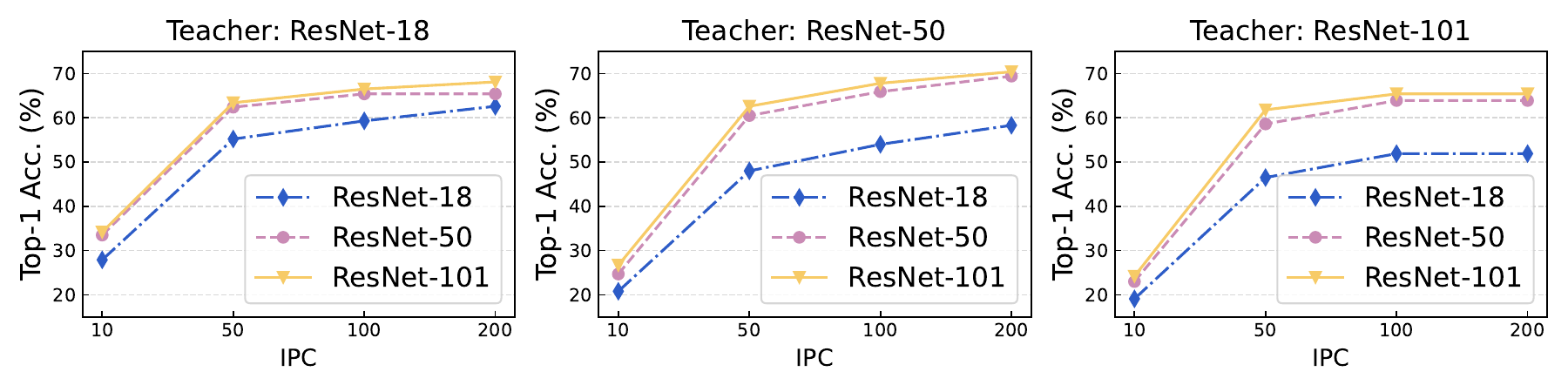}
  \caption{\textbf{Top-1 Accuracy$\uparrow$ of ImageNet-1K on various teacher-student pairs}. The result of each pair increases consistently with larger IPC}
  \label{fig:imgnt_fig1}
\end{figure*}

{\bf ImageNet-1K and Tiny-ImageNet}
The TTM strategy demonstrates remarkable efficacy in large-scale DD tasks as presented in \cref{tab:large_data}.
The effectiveness stems from its ability to improve the quality of the synthetic data rather than imitate the performance of the original data.
Consequently, it facilitates the processing of large-scale datasets with reduced computational complexity and memory demands.
In terms of accuracy, the proposed D$^4$M sets new benchmarks, achieving 66.5\% and 51.0\% with IPC-100 on ImageNet-1K and Tiny-ImageNet.
Notably, it replicates the full dataset performance with 81.2\% and 81.9\%, respectively.
Moreover, our approach significantly surpasses the leading data-matching method, SRe$^2$L, across both datasets. 
This superiority is attributed to the integration of multi-modal fusion embedding in D$^4$M.

Benefit to the architecture-free synthesis process, the datasets distilled by D$^4$M exhibit versatility.
To substantiate this characteristic, we extract 200 categories from the distilled ImageNet as the distilled Tiny-ImageNet in accordance with the predefined mapping \cite{Le2015TinyIV}.
The experimental outcomes of D$^4$M-G in \cref{tab:large_data} demonstrate that our method not only manifests a pronounced distillation effect but also retains the applicability inherent to the original dataset.

\subsection{Matching Strategy Analysis}
\begin{table}
\centering
\begin{tabular}{llll}
\toprule
Ablation & R18 & R50 & R101 \\
\midrule
 & \multicolumn{3}{c}{Teacher: R18} \\
\cmidrule{2-4}
w/ STM & 23.6 & 29.7 & 32.3 \\
w/o STM & \cellcolor{lightgray!30}\textbf{27.9(+4.3)} & \cellcolor{lightgray!30}\textbf{33.5(+3.8)} & \cellcolor{lightgray!30}\textbf{34.2(+1.9)} \\
\midrule
 & \multicolumn{3}{c}{Teacher: R50} \\
\cmidrule{2-4}
w/ STM & 15.8 & 20.6 & 22.3 \\
w/o STM & \cellcolor{lightgray!30}\textbf{20.7(+4.9)} & \cellcolor{lightgray!30}\textbf{24.7(+4.1)} & \cellcolor{lightgray!30}\textbf{26.7(+4.4)} \\
\midrule
 & \multicolumn{3}{c}{Teacher: R101} \\
\cmidrule{2-4}
w/ STM & 12.5 & 16.0 & 17.6 \\
w/o STM & \cellcolor{lightgray!30}\textbf{19.4(+6.9)} & \cellcolor{lightgray!30}\textbf{23.0(+7.0)} & \cellcolor{lightgray!30}\textbf{24.2(+6.6)} \\
\bottomrule
\end{tabular}
\caption{\textbf{Comparison of Top-1 Accuracy$\uparrow$ on different matching strategy}. We use the R18 as the distribution matching architecture. All methods are evaluated with IPC-10.}
\label{tab:exp_dm}
\end{table}

As mentioned in \cref{sec:rw}, the DD task often uses the STM strategy to generate images. 
In order to validate the superiority of TTM strategy, we conduct the comparative experiments listed in \cref{tab:exp_dm}.
We execute the synthesis process through BN distribution matching on images distilled via D$^4$M, resulting in distribution-matched synthetic images.

It is evident that the test performance with STM failed regardless of the chosen teacher network.
The images distilled via D$^4$M encapsulate not only the salient features of the original prototypes but also the text information of category labels.
Therefore, the network solely trained with the original images proves inadequate for effectively managing such fused multi-modal features.
Should the fused features be aligned with these networks, it would result in the disruption of the fused information, thereby diminishing the overall accuracy.
It is worth noting that D$^4$M potentially offers high-quality initialization for STM, as it synthesizes images with higher testing accuracy compared to those derived from random white noise initialization.

\subsection{Prototype Analysis}
\label{sec:prototype_exp}
To ascertain the critical role of prototypes in D$^4$M, we conduct an ablation study on the diffusion process with random initialization and prototype initialization.
The results listed in \cref{tab:prototype} demonstrate that the incorporation of a learned prototype markedly enhances the effectiveness of D$^4$M.
\begin{table}
\centering
\begin{tabular}{llll}
\toprule
Ablation & R18 & R50 & R101 \\
\midrule
 & \multicolumn{3}{c}{Dataset: ImageNet-1K} \\
 \cmidrule{2-4}
w/o PT & 15.6 & 20.7 & 20.6 \\
w/ PT & \cellcolor{lightgray!30}\textbf{27.9(+12.3)} & \cellcolor{lightgray!30}\textbf{33.5(+12.8)} & \cellcolor{lightgray!30}\textbf{34.2(+13.6)} \\
\midrule
 & \multicolumn{3}{c}{Dataset: Tiny-ImageNet} \\
 \cmidrule{2-4}
w/o PT & 30.5 & 35.6 & 37.3 \\
w/ PT & \cellcolor{lightgray!30}\textbf{46.2(+15.7)} & \cellcolor{lightgray!30}\textbf{51.8(+16.2)} & \cellcolor{lightgray!30}\textbf{51.0(+13.7)} \\
\bottomrule
\end{tabular}
\caption{\textbf{Comparison of Top-1 Accuracy$\uparrow$ on different initialization of diffusion process}. PT is the abbreviation of Prototype. All methods are evaluated with IPC-10.}
\label{tab:prototype}
\end{table}

\begin{figure}
    \centering
    \includegraphics[width=\linewidth]{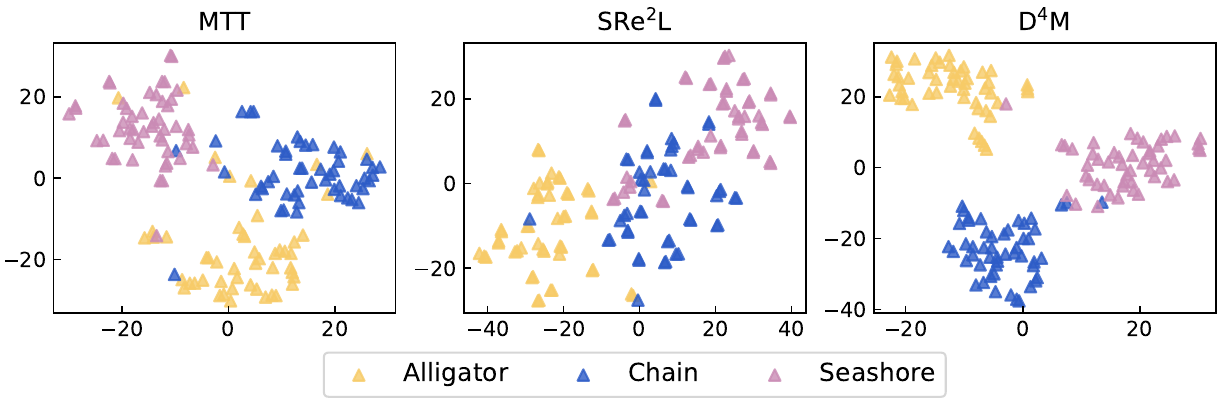}
    \caption{\textbf{T-SNE visualizations on Tiny-ImageNet}. The feature embedding distribution of D$^4$M displays more compact within classes and discriminative among classes.}
    \label{fig:tsne_tiny}
\end{figure}

To showcase the merits of the prototype intuitively, we employ ResNet-18 for feature extraction from the distilled dataset, followed by t-SNE for dimensionality reduction. 
The visualization results (\cref{fig:tsne_tiny}) reveal that the data synthesized via D$^4$M demonstrates enhanced inter-class discrimination and intra-class consistency.
% Therefore, D$^4$M attains superior Top-1 accuracy during training.

\subsection{Teacher-Student Network Analysis}
\label{sec:ts-anal}

\begin{table*}
\centering
\begin{tabular}{lccccc}
\toprule
\multirow{2}{*}{Teacher Network} & \multicolumn{5}{c}{Student Network} \\
\cmidrule{2-6}
 & ResNet-18  & MobileNet-V2 & EfficientNet-B0 & Swin-T & ViT-B \\
\midrule
ResNet-18 & 55.2 & 47.9 & \underline{55.4} & \textbf{58.1} & 45.5 \\
MobileNet-V2 & 47.6 & 42.9 & 49.8 & \textbf{58.9} & \underline{50.4} \\
% EfficientNet-B0 & 19.6 & 14.2 & \cellcolor{lightgray!30}\textbf{20.3} & 4.7 & 11.4 \\
Swin-T & 27.5 & 21.9 & 26.4 & \textbf{38.1} & \underline{34.2} \\
% ViT-B & \textbf{7.3} & 5.5 & 6.2 & 0.7 & \cellcolor{lightgray!30}5.0 \\
\bottomrule
\end{tabular}
\caption{\textbf{Top-1 Accuracy$\uparrow$ on ImageNet-1K with various teacher-student architectures}. ViT-based students show powerful learning ability with IPC-50.}
\label{tab:ts_cross50}
\end{table*}

We studied the performance of different teacher-student models with D$^4$M and the experimental results are shown in \cref{fig:imgnt_fig1}.
Under the same teacher network, the accuracy of ResNet-18, ResNet-50, and ResNet-101 increases gradually.
When IPC is small (such as 10 and 50), the student network trained with an enhanced teacher is prone to overfitting, resulting in reduced testing accuracy.
As IPC increases, the large network shows stronger learning ability and the Top-1 accuracy improves.
We further compare the performance of the distilled ImageNet on different teacher-student pairs, including CNNs and ViTs (\cref{tab:ts_cross50}).
As a student network, the ViT-based networks assimilate the inductive bias inherent in CNN-based teachers, leveraging its global attention mechanism to attain the best Top-1 accuracy.
Conversely, as a teacher network, ViT does not have such an inductive bias characteristic, yielding suboptimal results on their student networks.
Nevertheless, ViT-based students consistently achieve superior Top-1 accuracy.

\subsection{Qualitative Analysis}
A pivotal advantage of D$^4$M lies in its utilization of the outputs from the image decoder $\mathcal{D}$ as the distilled dataset, avoiding the need for STM. 
This implies that the pixel space of the generated image remains unaltered by any matching optimization, thereby preserving the reality of the distilled image. 
\Cref{fig:vis_1,fig:vis_2} exemplify the superior image quality achieved by D$^4$M in comparison to its counterparts.
It is evident that the D$^4$M method not only guarantees the high resolution of the distilled image and preserves the integrity of semantic information but also ensures the richness of features within the same category.
More visualizations and analysis can be found in supplementary material.

\begin{table}[h]
\centering
\begin{tabular}{lccc}
\toprule
Method & Resolution & Time(s)$\downarrow$ & GPU(GB)$\downarrow$ \\
\midrule
 & \multicolumn{3}{c}{Dataset: ImageNet-1K} \\
\cmidrule{2-4}
MTT$^\dagger$ & 128$\times$128 & 45.0 & 79.9 \\
TESLA$^\dagger$ & 64$\times$64 & 46.0 & 13.9 \\
SRe$^2$L & 224$\times$224 & 5.2 & 34.8 \\
D$^4$M & \cellcolor{lightgray!30}512$\times$512 & \cellcolor{lightgray!30}\textbf{2.7} & \cellcolor{lightgray!30}\textbf{6.1} \\
\midrule
 & \multicolumn{3}{c}{Dataset: Tiny-ImageNet} \\
\cmidrule{2-4}
MTT & 64$\times$64 & 5.4 & 48.9 \\
SRe$^2$L & 64$\times$64 & 11.0 & 33.8 \\
D$^4$M & \cellcolor{lightgray!30}512$\times$512 & \cellcolor{lightgray!30}\textbf{2.7} & \cellcolor{lightgray!30}\textbf{6.1} \\
\bottomrule
\end{tabular}
\caption{\textbf{Synthesis time$\downarrow$ and GPU memory$\downarrow$ cost on large-scale datasets}. $\dagger$: The runtime of MTT~\cite{cazenavette2022dataset} and TESLA~\cite{cui2023scaling} on ImageNet-1K are measured for 10 iterations (500 matching steps).}
\label{tab:time_gpu}
\end{table}

\subsection{Distillation Cost Analysis}
\begin{figure}[h]
  \centering
  \includegraphics[width=\linewidth]{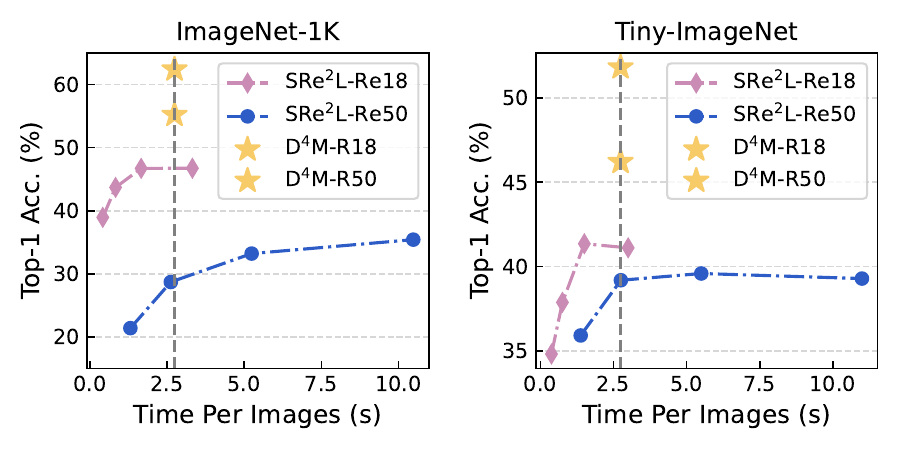}
  \caption{\textbf{Top-1 Accuracy$\uparrow$ and synthesis time$\downarrow$ on large-scale datasets}. D$^4$M is architecture-free at synthesis time, thereby a constant runtime cost. Re is the abbreviation of Recover.}
  \label{fig:time_acc_fig}
\end{figure}

We conduct the analysis of GPU memory consumption across various DD methods, with the corresponding results presented in \cref{tab:time_gpu}.
Notably, the architecture-free nature of D$^4$M during synthesis ensures the fixed time and GPU memory costs.
When considering STM and DTM, we observe an increase in both time and GPU memory usage with the enlargement of the matching architecture.
For instance, the peak GPU memory utilization for SRe$^2$L in the recovery of a 64$\times$64 image on ConvNet is 4.2 GB, whereas on ResNet-50, it reaches a substantial 33.8 GB. 
Similarly, when synthesizing a 64$\times$64 image on ConvNet, MTT demands a peak GPU memory of 48.9 GB.
Furthermore, the number of iteration steps impacts the generation time for a single image in data matching. 
With the increased iteration steps, the time cost for SRe$^2$L to recover a 224$\times$224 image on ResNet-50 gradually rises from 1.31s to 10.48s. 
Notably, D$^4$M demonstrates a remarkable reduction in time cost by a factor of 3.82 when compared to SRe$^2$L.
\Cref{fig:time_acc_fig} reveals that D$^4$M attains best accuracy at a constant time cost.

% \begin{table*}
% \centering
% \begin{tabular}{lccccccc}
% \toprule
% Dataset & IPC & SRe$^2$L(R18) & D$^4$M(R18) & SRe$^2$L(R50) & D$^4$M(R50) & SRe$^2$L(R101) & D$^4$M(R101) \\
% \midrule
% \multirow{4}{*}{ImageNet-1K} & 10 & 21.3 & \cellcolor{lightgray!30}\textbf{27.9} & 28.4 & \cellcolor{lightgray!30}\textbf{33.5} & 30.9 & \cellcolor{lightgray!30}\textbf{34.2} \\
%  & 50 & 46.8 & \cellcolor{lightgray!30}\textbf{55.2} & 55.6 & \cellcolor{lightgray!30}\textbf{62.4} & 60.8 & \cellcolor{lightgray!30}\textbf{63.4} \\
%  & 100 & 52.8 & \cellcolor{lightgray!30}\textbf{59.3} & 61.0 & \cellcolor{lightgray!30}\textbf{65.4} & 65.8 & \cellcolor{lightgray!30}\textbf{66.5} \\
%  & 200 & 57.0 & \cellcolor{lightgray!30} & 64.4 & \cellcolor{lightgray!30} & 65.9 & \cellcolor{lightgray!30} \\
% \bottomrule
% \end{tabular}
% \caption{test}
% \label{tab:main_exp}
% \end{table*}

%% file: sec/5_conclusion.tex
\section{Conclusion}
We introduce D$^4$M, a novel and efficient dataset distillation framework leveraging the TTM strategy. 
For the first time, D$^4$M addresses the cross-architecture generalization issue by integrating the principles of diffusion models with prototype learning.
The distilled dataset not only boasts realistic and high-resolution images with limited resources but also exhibits a versatility comparable to that of the full dataset.
D$^4$M demonstrates outstanding performance compared to other dataset distillation methods, particularly when applied to large-scale datasets such as ImageNet-1K.
Last but not least, rethinking the relationship between generative models and dataset distillation offers fresh perspectives, paving the way for the community to develop more efficient dataset distillation methods in future endeavors.

\noindent \textbf{Limitation and future works.} 
In the situation of extreme distillation (IPC-1/10), we observe a significant performance degradation. Our future work will concentrate on refining the distillation process for this challenging scenario and try to distill more real-world multi-modal datasets.

\noindent \textbf{Acknowledgement.} This work is supported by the National Natural Science Foundation of China (No. 12071458).

%% file: sec/X_suppl.tex
\clearpage
\maketitlesupplementary
\setcounter{section}{0}
\section{Experimental Settings}
\label{sec:exp_set}

In our experimental framework, we primarily concentrate on the parameters of the synthesis and the Training-Time Matching (TTM) processes. 
For the synthesis phase, Stable Diffusion (V1-5) serves as the core mechanism in Latent Diffusion Model implementation. 
Based on the insights of \cref{sec:sens_anal}, we calibrate the \textit{strength} and \textit{guidance scale} parameters at 0.7 and 8, respectively. 
During the prototype learning, the Mini-Batch $k$-Means algorithm is employed, with an in-depth ablation study of cluster number variations presented in \cref{sec:n_pros}. 
Furthermore, in scenarios where the IPC is less than 100, we adjust the cluster numbers to match the IPC. 
Within the TTM process, the comprehensive parameter settings of student networks are provided in \cref{tab:conf_student}.

\section{Hyper-parameter Analysis}
\subsection{Sensitivity Analysis}
\label{sec:sens_anal}
\begin{figure}[ht]
  \centering
  \begin{subfigure}{\linewidth}
    \includegraphics[width=\textwidth]{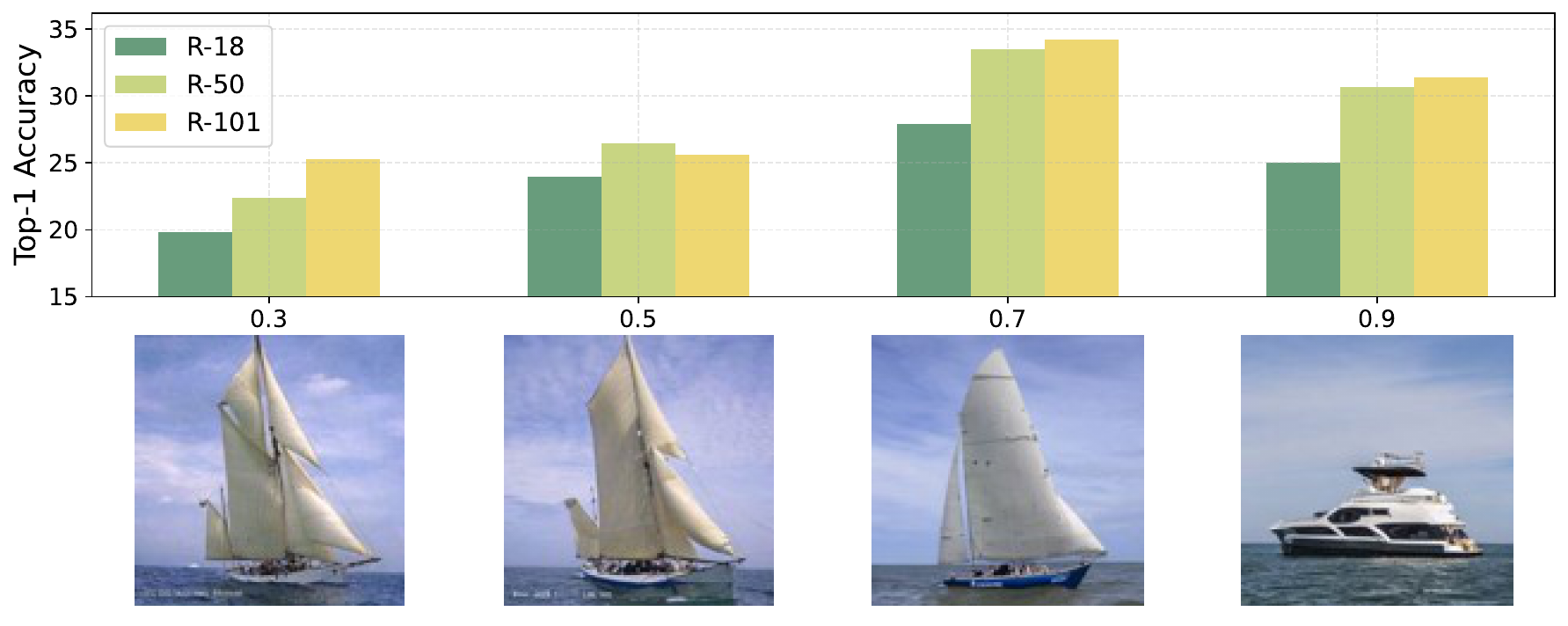}
    \caption{Sensitivity Analysis of \textit{strength} ($\textit{guidance scale}=8$)}
    \label{fig:strength}
  \end{subfigure}
  \hfill
  \begin{subfigure}{\linewidth}
    \includegraphics[width=\textwidth]{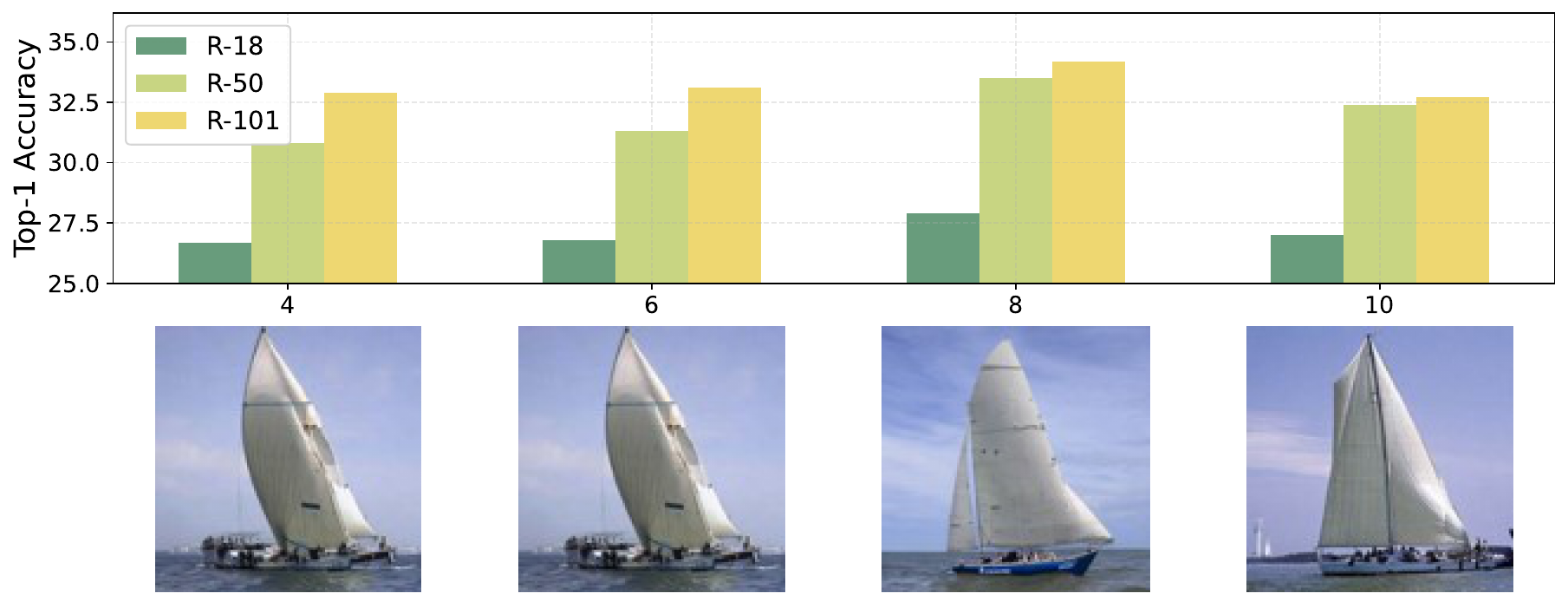}
    \caption{Sensitivity Analysis of \textit{guidance scale} ($\textit{strength}=0.7$)}
    \label{fig:scale}
  \end{subfigure}
  \caption{\textbf{Sensitivity analysis of \textit{strength} and \textit{guidance scale}}. Quantitative results are evaluated on ResNet. Furthermore, qualitative results are presented to illustrate the variations corresponding to parameter adjustments.}
  \label{fig:hyper-paras}
\end{figure}

There are two hyper-parameters in the diffusion model with text prompts, \ie \textit{strength} ($0 < s < 1$) and \textit{guidance scale} ($g>1$).
Conceptually, the \textit{strength} quantifies the extent of noise infusion into the latent features (prototypes). 
The diffusion model predominantly disregards these features in scenarios where \textit{strength} equals 1. 
Furthermore, an elevated \textit{guidance scale} fosters the generation of images that more precisely align with the text prompt.
Based on the hyper-parameter tuning results in \cref{fig:strength} and \cref{fig:scale}, we suggest setting $\textit{strength}=0.7$ and $\textit{guidance scale}=8$.

\subsection{Number of Prototypes}
\label{sec:n_pros}

To ensure the feature diversity of the distilled dataset, multiple prototypes are learned for each category in our experiments. 
We select 10 or 50 prototypes to generate distilled ImageNet-1K datasets (IPC-100/200) respectively, \ie synthesizing multiple images per prototype.
These datasets are then trained across three distinct ResNet architectures, with the corresponding outcomes detailed in \cref{tab:exp_prototype}.

\begin{figure*}
  \centering
  \includegraphics[width=\linewidth]{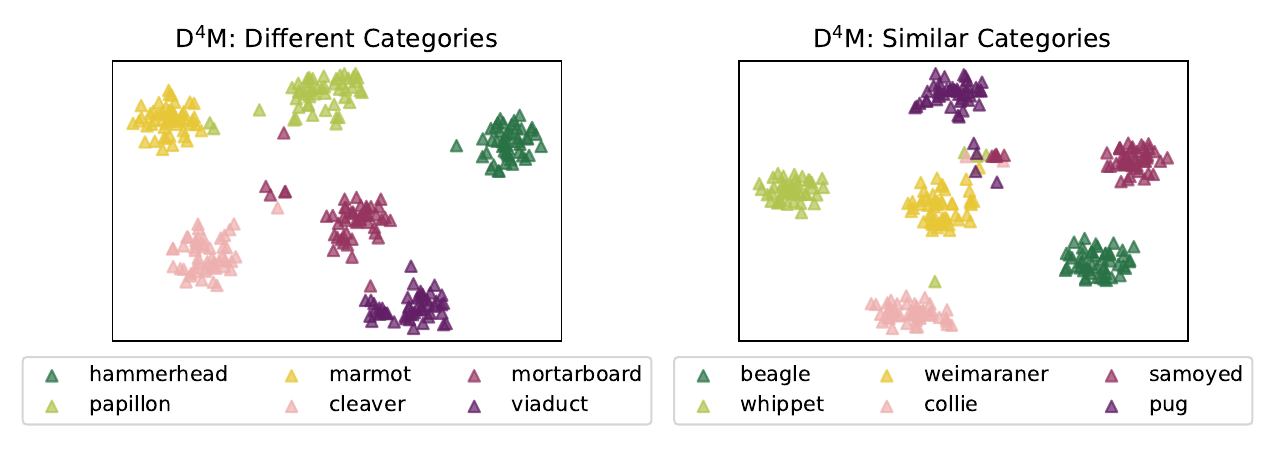}
  \caption{\textbf{T-SNE visualizations on ImageNet-1K.} The features are extracted by ResNet-18.}
  \label{fig:tsne_imagenet}
\end{figure*}

\begin{table}[ht]
\centering
\begin{tabular}{cccccc}
\toprule
\cellcolor{lightgray!30}IPC & \cellcolor{lightgray!30}Prototypes & \cellcolor{lightgray!30}R18 & \cellcolor{lightgray!30}R50 & \cellcolor{lightgray!30}R101 & \cellcolor{lightgray!30}p-value \\
\midrule
\multirow{2}{*}{100} & 10 & 59.0 & 64.4 & 65.9 & \multirow{4}{*}{0.7391} \\
 & 50 & 59.3 & 65.4 & 66.5 & \\
\cmidrule{1-5}
\multirow{2}{*}{200} & 10 & 62.4 & 67.6 & 68.2 & \\
 & 50 & 62.6 & 67.8 & 68.1 & \\
\bottomrule
\end{tabular}
\caption{\textbf{Ablation study on the cluster numbers.} We utilize 10 or 50 prototypes to synthesize images for IPC-100/200 and evaluate them with ResNet.}\label{tab:exp_prototype}
\end{table}

\begin{table*}
    \centering
    \begin{subtable}[b]{0.43\textwidth}
        \centering
        \caption{ImageNet-1K and Tiny-ImageNet}
        \begin{tabular}{cc}
            \toprule
            \cellcolor{lightgray!30}Settings & \cellcolor{lightgray!30}Values \\
            \midrule
            network & ResNet \\
            input size & 224 \\
            batch size & 1024 \\
            epoch & 300 \\
            augmentation & RandomResizedCrop \\
            min scale  & 0.08 \\
            max scale  & 1 \\
            temperature & 20 \\
            optimizer & AdamW \\
            learning rate & 0.001 \\
            weight decay & 0.01 \\
            learning rate schedule  & cosine decay \\
            \bottomrule
        \end{tabular}
    \end{subtable}
    % \begin{subtable}[b]{0.43\textwidth}
    %     \centering
    %     \caption{Tiny-ImageNet Student}
    %     \begin{tabular}{cc}
    %         \toprule
    %         \cellcolor{lightgray!30}Settings & \cellcolor{lightgray!30}Values \\
    %         \midrule
    %         network & ResNet \\
    %         input size & 224 \\
    %         batch size & 1024 \\
    %         epoch & 300 \\
    %         augmentation & RandomResizedCrop \\
    %         min scale  & 0.08 \\
    %         max scale  & 1 \\
    %         temperature & 20 \\
    %         optimizer & AdamW \\
    %         learning rate & 0.001 \\
    %         weight decay & 0.01 \\
    %         learning rate schedule  & cosine decay \\
    %         \bottomrule
    %     \end{tabular}
    % \end{subtable}
    \begin{subtable}[b]{0.43\textwidth}
        \centering
        \caption{CIFAR-10 and CIFAR-100}
        \begin{tabular}{cc}
            \toprule
            \cellcolor{lightgray!30}Settings & \cellcolor{lightgray!30}Values \\
            \midrule
            network & ConvNet \\
            input size & 32 \\
            batch size & 100 \\
            epoch & 500 \\
            augmentation & RandomResizedCrop \\
            min scale  & 0.08 \\
            max scale  & 1 \\
            temperature & 20 \\
            optimizer & AdamW \\
            learning rate & 0.001 \\
            weight decay & 0.01 \\
            learning rate schedule  & cosine decay \\
            \bottomrule
        \end{tabular}
    \end{subtable}
    % \begin{subtable}[b]{0.43\textwidth}
    %     \centering
    %     \caption{CIFAR-10 Student}
    %     \begin{tabular}{cc}
    %         \toprule
    %         \cellcolor{lightgray!30}Settings & \cellcolor{lightgray!30}Values \\
    %         \midrule
    %         network & ConvNet \\
    %         input size & 32 \\
    %         batch size & 100 \\
    %         epoch & 500 \\
    %         augmentation & RandomResizedCrop \\
    %         min scale  & 0.08 \\
    %         max scale  & 1 \\
    %         temperature & 20 \\
    %         optimizer & AdamW \\
    %         learning rate & 0.001 \\
    %         weight decay & 0.01 \\
    %         learning rate schedule  & cosine decay \\
    %         \bottomrule
    %     \end{tabular}
    % \end{subtable}   
\caption{Parameter settings of the student networks.}
\label{tab:conf_student}
\end{table*}

Given the marginal disparity observed between the experimental results of the two groups, we conducted an independent sample t-test.
The alternative hypothesis is that the true difference in means is not equal to 0.
According to the p-value, at a significance threshold of 0.05, the performance variations of each group are not statistically significant, which means that the distilled datasets are not sensitive to the number of prototypes.

In addition, the t-SNE visualization results of D$^4$M on ImageNet-1K are displayed in \cref{fig:tsne_imagenet}.
Except for a few outliers, the features extracted from the D$^4$M distilled ImageNet-1K dataset are compact and discriminative for both different and similar categories.

\section{Quantitative Analysis}

% In the main text, we explore the improvement of input-output space consistency constraints in solving cross-architecture generalization problems.
% Here we conduct a direct comparison between the image quality resulting from our distillation method and the established benchmarks, as delineated in \cref{tab:quality}.

In the main text, we delve into the enhancement of input-output image space consistency constraints for addressing cross-architecture generalization challenges. 
This section presents a direct comparative analysis of the image quality yielded by D$^4$M against the benchmark, as detailed in \cref{tab:quality}.

Firstly, we employ the Inception Score (IS) to assess the 
clarity $p(y \mid {x})$ of the synthetic images and the feature diversity $p(y)$ of the generative model $G$.
The IS quantifies the KL divergence between the probability distribution and the conditional probability distribution of the features, as extracted by the Inception V3 model:
\begin{equation}
\mathrm{IS}=\exp \left(\mathbb{E}_{{x} \sim p_G} D_{KL}(p(y \mid {x}) \| p(y))\right).
\end{equation}
Moreover, to demonstrate that the D$^4$M enhances the consistency between synthetic and real images, we compute the Fr\'{e}chet Inception Distance (FID) and Kernel Inception Distance (KID) metrics for these datasets.
Empirical evaluations demonstrate that D$^4$M is capable of generating a variety of high-resolution images while maintaining consistency between the input and output image spaces.

\begin{table}[!ht]
\centering
\begin{tabular}{llccc}
\toprule
Dataset & Method & IS$\uparrow$ & FID$\downarrow$ & KID$\downarrow$ \\
\midrule
\multirow{2}{*}{ImageNet-1K} & SRe$^2$L & 28.872 & 59.119 & 0.047 \\
& D$^4$M & \cellcolor{lightgray!30}\textbf{49.381} & \cellcolor{lightgray!30}\textbf{9.419} & \cellcolor{lightgray!30}\textbf{0.003} \\
\midrule
\multirow{2}{*}{Tiny-ImageNet} & SRe$^2$L & 6.243 & 74.814 & 0.055 \\
& D$^4$M & \cellcolor{lightgray!30}\textbf{25.866} & \cellcolor{lightgray!30}\textbf{34.702} & \cellcolor{lightgray!30}\textbf{0.020} \\
\bottomrule
\end{tabular}
\caption{Quantitative results of distilled image. Comparing the quality of distilled images using IPC-50 on ImageNet-1K and Tiny-ImageNet, D$^4$M consistently outperforms SRe$^2$L across IS, FID, and KID metrics. This demonstrates that the distilled images produced by D$^4$M exhibit higher image quality.}
\label{tab:quality}
\end{table}

\section{More Visualizations}

\begin{figure*}
\centering
\setlength{\tabcolsep}{0pt}
\renewcommand{\arraystretch}{0}
\begin{tabular}{llllllllll}
\includegraphics[width=0.09\linewidth]{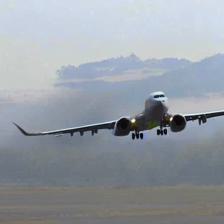} & 
\includegraphics[width=0.09\linewidth]{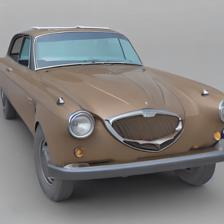} & 
\includegraphics[width=0.09\linewidth]{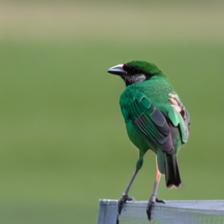} & 
\includegraphics[width=0.09\linewidth]{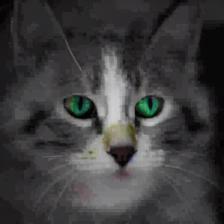} & 
\includegraphics[width=0.09\linewidth]{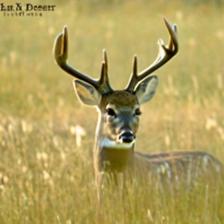} & 
\includegraphics[width=0.09\linewidth]{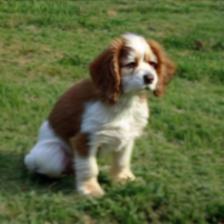} & 
\includegraphics[width=0.09\linewidth]{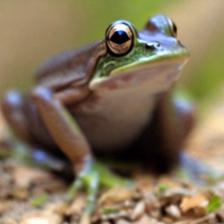} & 
\includegraphics[width=0.09\linewidth]{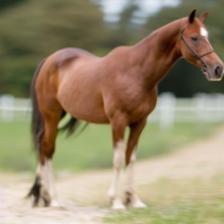} & 
\includegraphics[width=0.09\linewidth]{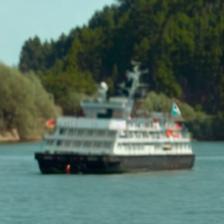} & 
\includegraphics[width=0.09\linewidth]{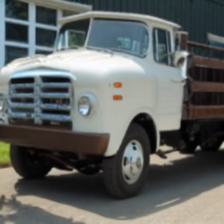} \\
\end{tabular}
\caption{More visualizations selected from the distilled CIFAR-10 (Class 0-9)}
\label{fig:vis_cifar10}
\end{figure*}

\begin{figure*}
\centering
\setlength{\tabcolsep}{0pt}
\renewcommand{\arraystretch}{0}
\begin{tabular}{llllllllll}
\includegraphics[width=0.09\linewidth]{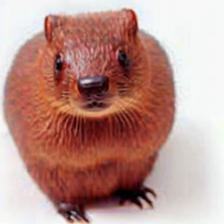} & 
\includegraphics[width=0.09\linewidth]{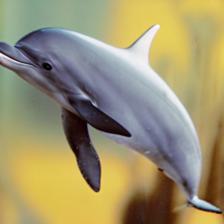} & 
\includegraphics[width=0.09\linewidth]{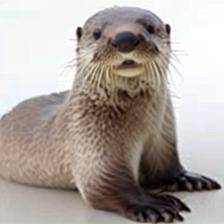} & 
\includegraphics[width=0.09\linewidth]{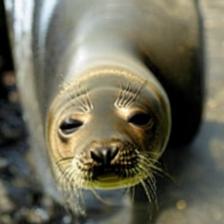} & 
\includegraphics[width=0.09\linewidth]{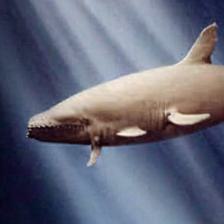} & 
\includegraphics[width=0.09\linewidth]{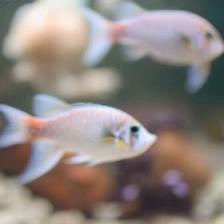} & 
\includegraphics[width=0.09\linewidth]{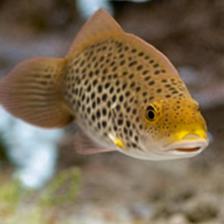} & 
\includegraphics[width=0.09\linewidth]{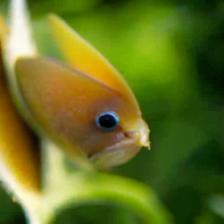} & 
\includegraphics[width=0.09\linewidth]{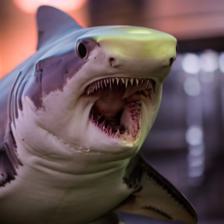} & 
\includegraphics[width=0.09\linewidth]{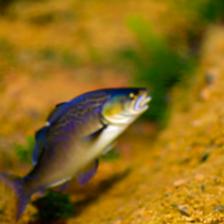} \\
\includegraphics[width=0.09\linewidth]{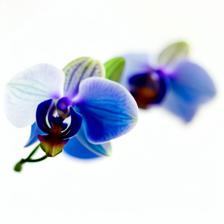} & 
\includegraphics[width=0.09\linewidth]{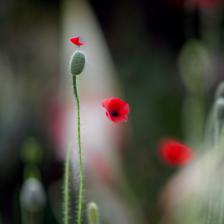} & 
\includegraphics[width=0.09\linewidth]{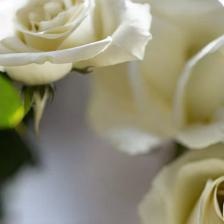} & 
\includegraphics[width=0.09\linewidth]{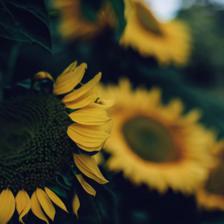} & 
\includegraphics[width=0.09\linewidth]{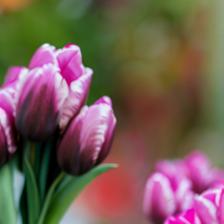} & 
\includegraphics[width=0.09\linewidth]{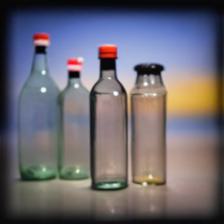} & 
\includegraphics[width=0.09\linewidth]{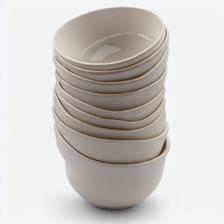} & 
\includegraphics[width=0.09\linewidth]{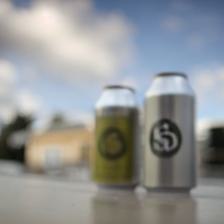} & 
\includegraphics[width=0.09\linewidth]{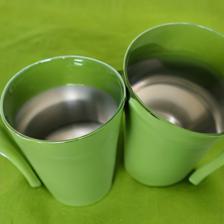} & 
\includegraphics[width=0.09\linewidth]{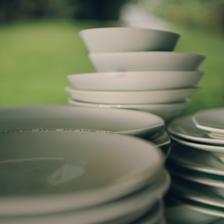} \\
\end{tabular}
\caption{More visualizations selected from the distilled CIFAR-100 (Class 0-99)}
\label{fig:vis_cifar100}
\end{figure*}

We randomly select the visualizations to enhance the understanding of our methods and easier to reference.
The distilled CIFAR-10 and CIFAR-100 are shown in \cref{fig:vis_cifar10} and \cref{fig:vis_cifar100}.
% Furthermore, the distilled ImageNet-1K is shown in \cref{fig:vis1}-\cref{fig:vis10}. 
(\textbf{more pages after this paragraph})

\begin{figure*}
\centering
\setlength{\tabcolsep}{0pt}
\renewcommand{\arraystretch}{0}
\begin{tabular}{llllllllll}
\includegraphics[width=0.09\linewidth]{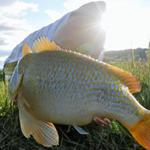} & \includegraphics[width=0.09\linewidth]{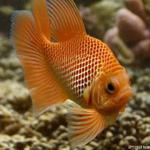} & 
\includegraphics[width=0.09\linewidth]{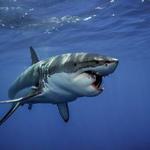} & \includegraphics[width=0.09\linewidth]{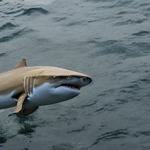} & \includegraphics[width=0.09\linewidth]{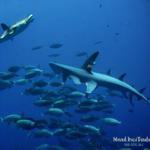} & \includegraphics[width=0.09\linewidth]{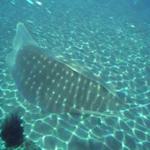} & \includegraphics[width=0.09\linewidth]{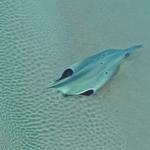} & \includegraphics[width=0.09\linewidth]{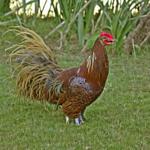} & \includegraphics[width=0.09\linewidth]{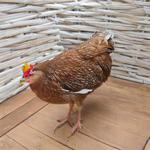} & \includegraphics[width=0.09\linewidth]{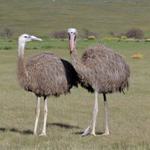} \\
\includegraphics[width=0.09\linewidth]{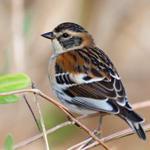} & \includegraphics[width=0.09\linewidth]{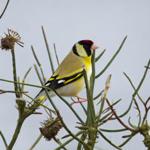} & 
\includegraphics[width=0.09\linewidth]{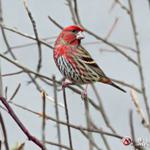} & \includegraphics[width=0.09\linewidth]{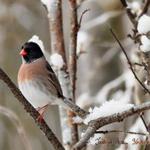} & \includegraphics[width=0.09\linewidth]{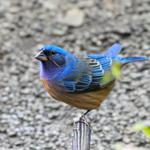} & \includegraphics[width=0.09\linewidth]{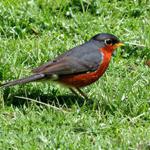} & \includegraphics[width=0.09\linewidth]{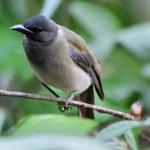} & \includegraphics[width=0.09\linewidth]{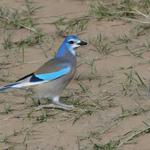} & \includegraphics[width=0.09\linewidth]{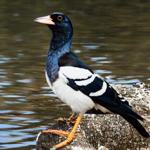} & \includegraphics[width=0.09\linewidth]{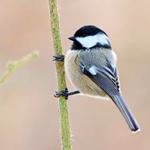} \\
\end{tabular}
\caption{More visualizations selected from the distilled ImageNet-1K (Class 0-99)}
\label{fig:vis1}
\end{figure*}

\begin{figure*}
\centering
\setlength{\tabcolsep}{0pt}
\renewcommand{\arraystretch}{0}
\begin{tabular}{llllllllll}
\includegraphics[width=0.09\linewidth]{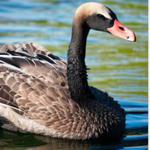} & \includegraphics[width=0.09\linewidth]{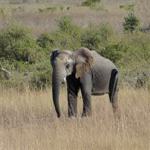} & 
\includegraphics[width=0.09\linewidth]{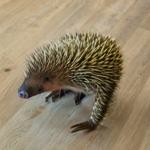} & \includegraphics[width=0.09\linewidth]{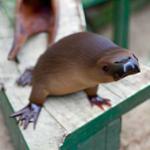} & \includegraphics[width=0.09\linewidth]{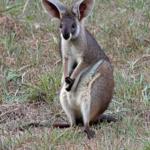} & \includegraphics[width=0.09\linewidth]{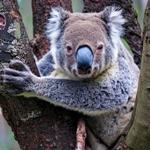} & \includegraphics[width=0.09\linewidth]{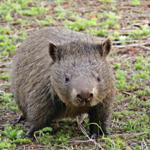} & \includegraphics[width=0.09\linewidth]{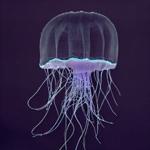} & \includegraphics[width=0.09\linewidth]{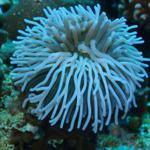} & \includegraphics[width=0.09\linewidth]{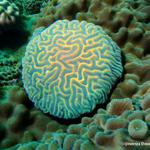} \\
\includegraphics[width=0.09\linewidth]{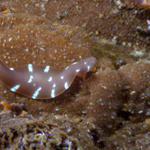} & \includegraphics[width=0.09\linewidth]{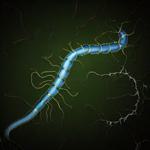} & 
\includegraphics[width=0.09\linewidth]{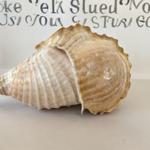} & \includegraphics[width=0.09\linewidth]{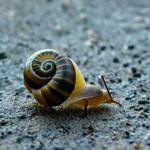} & \includegraphics[width=0.09\linewidth]{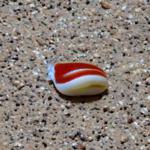} & \includegraphics[width=0.09\linewidth]{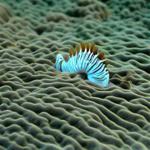} & \includegraphics[width=0.09\linewidth]{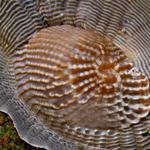} & \includegraphics[width=0.09\linewidth]{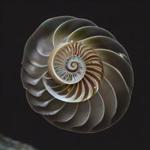} & \includegraphics[width=0.09\linewidth]{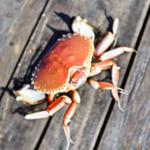} & \includegraphics[width=0.09\linewidth]{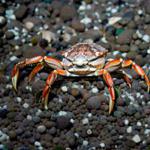} \\
\end{tabular}
\caption{More visualizations selected from the distilled ImageNet-1K (Class 100-199)}
\label{fig:vis2}
\end{figure*}

\begin{figure*}
\centering
\setlength{\tabcolsep}{0pt}
\renewcommand{\arraystretch}{0}
\begin{tabular}{llllllllll}
\includegraphics[width=0.09\linewidth]{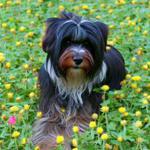} & \includegraphics[width=0.09\linewidth]{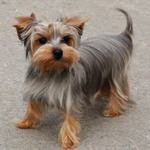} & 
\includegraphics[width=0.09\linewidth]{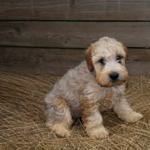} & \includegraphics[width=0.09\linewidth]{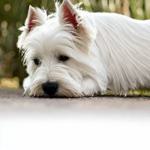} & \includegraphics[width=0.09\linewidth]{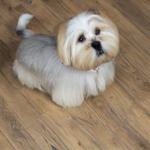} & \includegraphics[width=0.09\linewidth]{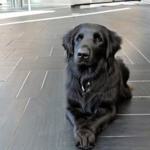} & \includegraphics[width=0.09\linewidth]{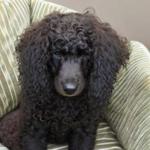} & \includegraphics[width=0.09\linewidth]{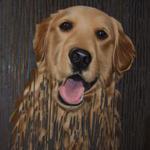} & \includegraphics[width=0.09\linewidth]{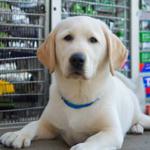} & \includegraphics[width=0.09\linewidth]{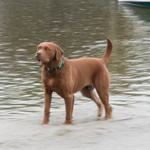} \\
\includegraphics[width=0.09\linewidth]{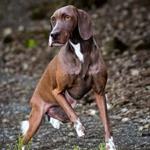} & \includegraphics[width=0.09\linewidth]{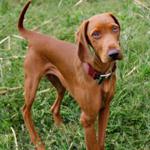} & 
\includegraphics[width=0.09\linewidth]{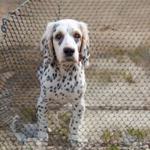} & \includegraphics[width=0.09\linewidth]{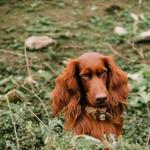} & \includegraphics[width=0.09\linewidth]{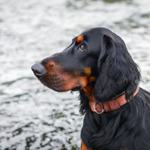} & \includegraphics[width=0.09\linewidth]{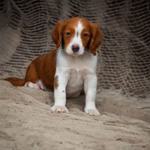} & \includegraphics[width=0.09\linewidth]{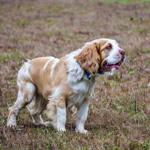} & \includegraphics[width=0.09\linewidth]{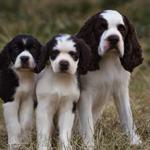} & \includegraphics[width=0.09\linewidth]{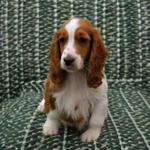} & \includegraphics[width=0.09\linewidth]{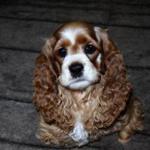} \\
\end{tabular}
\caption{More visualizations selected from the distilled ImageNet-1K (Class 200-299)}
\label{fig:vis3}
\end{figure*}

\begin{figure*}
\centering
\setlength{\tabcolsep}{0pt}
\renewcommand{\arraystretch}{0}
\begin{tabular}{llllllllll}
\includegraphics[width=0.09\linewidth]{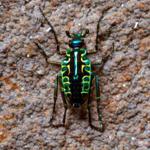} & \includegraphics[width=0.09\linewidth]{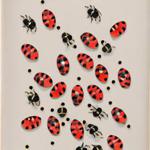} & 
\includegraphics[width=0.09\linewidth]{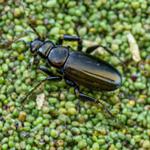} & \includegraphics[width=0.09\linewidth]{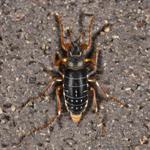} & \includegraphics[width=0.09\linewidth]{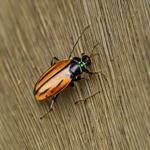} & \includegraphics[width=0.09\linewidth]{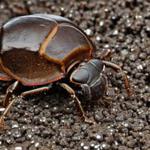} & \includegraphics[width=0.09\linewidth]{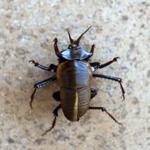} & \includegraphics[width=0.09\linewidth]{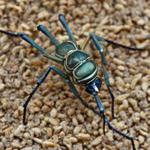} & \includegraphics[width=0.09\linewidth]{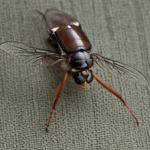} & \includegraphics[width=0.09\linewidth]{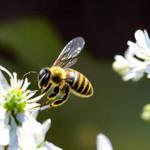} \\
\includegraphics[width=0.09\linewidth]{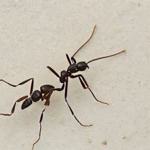} & \includegraphics[width=0.09\linewidth]{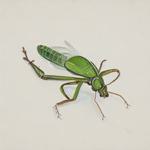} & 
\includegraphics[width=0.09\linewidth]{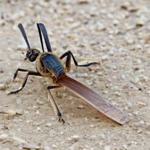} & \includegraphics[width=0.09\linewidth]{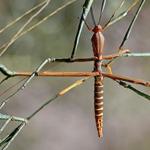} & \includegraphics[width=0.09\linewidth]{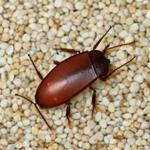} & \includegraphics[width=0.09\linewidth]{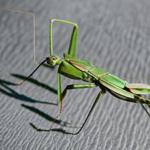} & \includegraphics[width=0.09\linewidth]{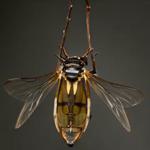} & \includegraphics[width=0.09\linewidth]{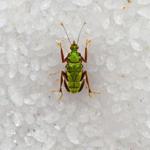} & \includegraphics[width=0.09\linewidth]{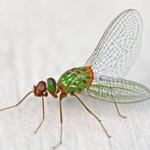} & \includegraphics[width=0.09\linewidth]{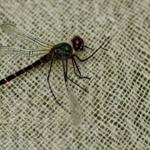} \\
\end{tabular}
\caption{More visualizations selected from the distilled ImageNet-1K (Class 300-399)}
\label{fig:vis4}
\end{figure*}

\begin{figure*}
\centering
\setlength{\tabcolsep}{0pt}
\renewcommand{\arraystretch}{0}
\begin{tabular}{llllllllll}
\includegraphics[width=0.09\linewidth]{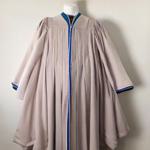} & \includegraphics[width=0.09\linewidth]{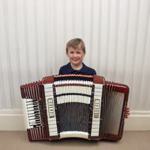} & 
\includegraphics[width=0.09\linewidth]{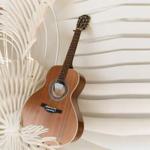} & \includegraphics[width=0.09\linewidth]{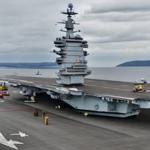} & \includegraphics[width=0.09\linewidth]{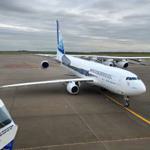} & \includegraphics[width=0.09\linewidth]{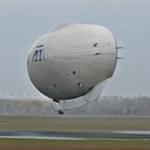} & \includegraphics[width=0.09\linewidth]{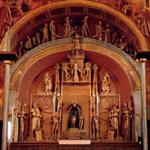} & \includegraphics[width=0.09\linewidth]{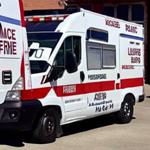} & \includegraphics[width=0.09\linewidth]{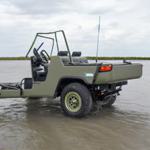} & \includegraphics[width=0.09\linewidth]{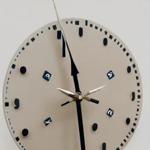} \\
\includegraphics[width=0.09\linewidth]{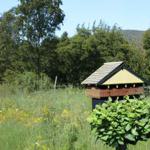} & \includegraphics[width=0.09\linewidth]{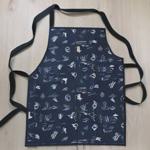} & 
\includegraphics[width=0.09\linewidth]{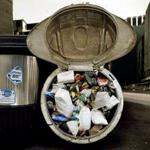} & \includegraphics[width=0.09\linewidth]{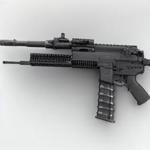} & \includegraphics[width=0.09\linewidth]{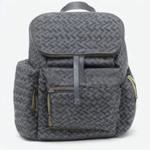} & \includegraphics[width=0.09\linewidth]{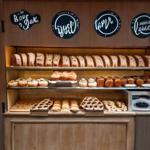} & \includegraphics[width=0.09\linewidth]{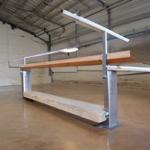} & \includegraphics[width=0.09\linewidth]{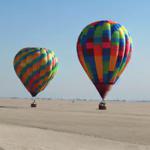} & \includegraphics[width=0.09\linewidth]{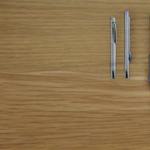} & \includegraphics[width=0.09\linewidth]{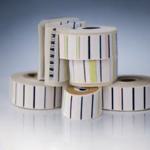} \\
\end{tabular}
\caption{More visualizations selected from the distilled ImageNet-1K (Class 400-499)}
\label{fig:vis5}
\end{figure*}

\begin{figure*}
\centering
\setlength{\tabcolsep}{0pt}
\renewcommand{\arraystretch}{0}
\begin{tabular}{llllllllll}
\includegraphics[width=0.09\linewidth]{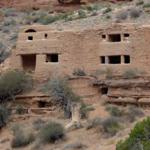} & \includegraphics[width=0.09\linewidth]{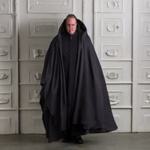} & 
\includegraphics[width=0.09\linewidth]{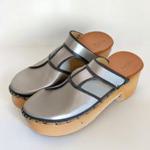} & \includegraphics[width=0.09\linewidth]{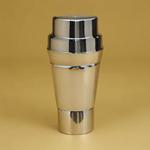} & \includegraphics[width=0.09\linewidth]{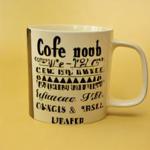} & \includegraphics[width=0.09\linewidth]{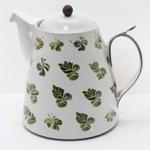} & \includegraphics[width=0.09\linewidth]{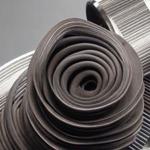} & \includegraphics[width=0.09\linewidth]{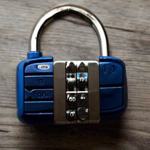} & \includegraphics[width=0.09\linewidth]{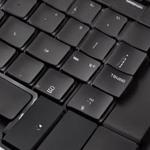} & \includegraphics[width=0.09\linewidth]{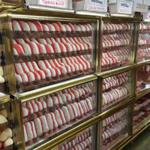} \\
\includegraphics[width=0.09\linewidth]{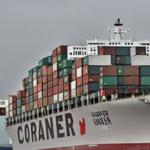} & \includegraphics[width=0.09\linewidth]{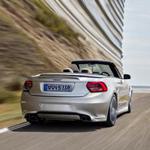} & 
\includegraphics[width=0.09\linewidth]{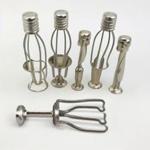} & \includegraphics[width=0.09\linewidth]{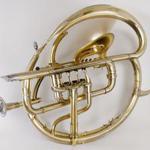} & \includegraphics[width=0.09\linewidth]{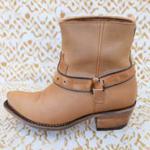} & \includegraphics[width=0.09\linewidth]{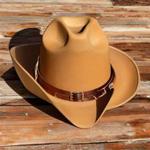} & \includegraphics[width=0.09\linewidth]{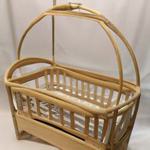} & \includegraphics[width=0.09\linewidth]{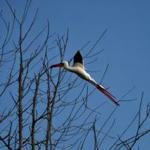} & \includegraphics[width=0.09\linewidth]{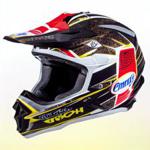} & \includegraphics[width=0.09\linewidth]{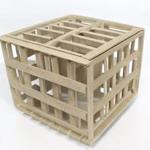} \\
\end{tabular}
\caption{More visualizations selected from the distilled ImageNet-1K (Class 500-599)}
\label{fig:vis6}
\end{figure*}

\begin{figure*}
\centering
\setlength{\tabcolsep}{0pt}
\renewcommand{\arraystretch}{0}
\begin{tabular}{llllllllll}
\includegraphics[width=0.09\linewidth]{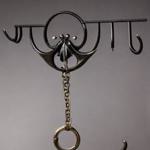} & \includegraphics[width=0.09\linewidth]{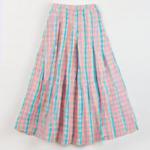} & 
\includegraphics[width=0.09\linewidth]{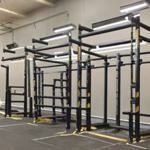} & \includegraphics[width=0.09\linewidth]{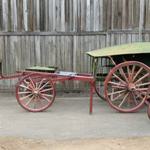} & \includegraphics[width=0.09\linewidth]{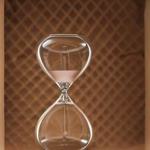} & \includegraphics[width=0.09\linewidth]{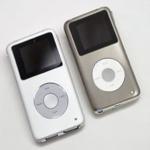} & \includegraphics[width=0.09\linewidth]{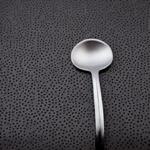} & \includegraphics[width=0.09\linewidth]{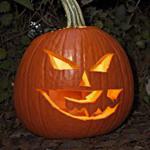} & \includegraphics[width=0.09\linewidth]{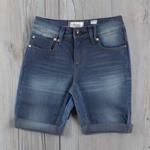} & \includegraphics[width=0.09\linewidth]{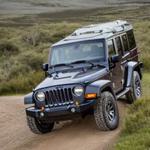} \\
\includegraphics[width=0.09\linewidth]{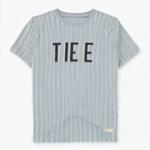} & \includegraphics[width=0.09\linewidth]{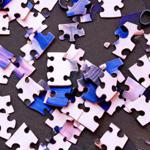} & 
\includegraphics[width=0.09\linewidth]{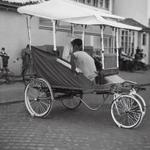} & \includegraphics[width=0.09\linewidth]{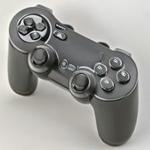} & \includegraphics[width=0.09\linewidth]{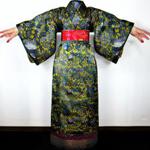} & \includegraphics[width=0.09\linewidth]{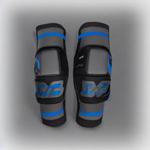} & \includegraphics[width=0.09\linewidth]{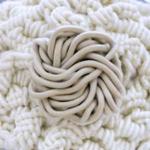} & \includegraphics[width=0.09\linewidth]{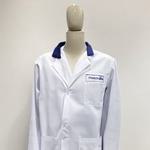} & \includegraphics[width=0.09\linewidth]{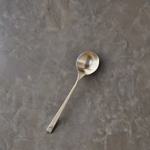} & \includegraphics[width=0.09\linewidth]{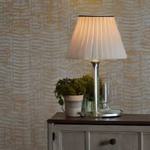} \\
\end{tabular}
\caption{More visualizations selected from the distilled ImageNet-1K (Class 600-699)}
\label{fig:vis7}
\end{figure*}

\begin{figure*}
\centering
\setlength{\tabcolsep}{0pt}
\renewcommand{\arraystretch}{0}
\begin{tabular}{llllllllll}
\includegraphics[width=0.09\linewidth]{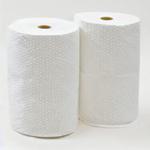} & \includegraphics[width=0.09\linewidth]{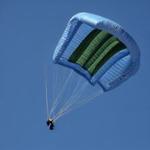} & 
\includegraphics[width=0.09\linewidth]{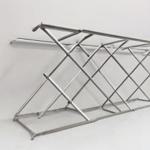} & \includegraphics[width=0.09\linewidth]{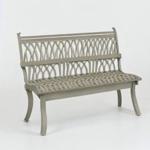} & \includegraphics[width=0.09\linewidth]{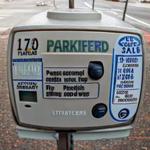} & \includegraphics[width=0.09\linewidth]{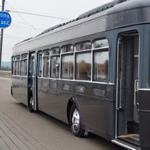} & \includegraphics[width=0.09\linewidth]{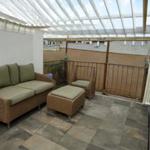} & \includegraphics[width=0.09\linewidth]{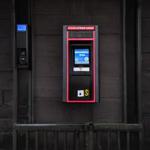} & \includegraphics[width=0.09\linewidth]{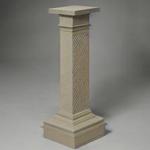} & \includegraphics[width=0.09\linewidth]{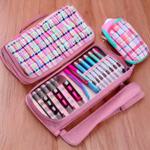} \\
\includegraphics[width=0.09\linewidth]{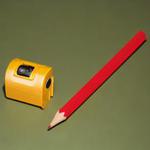} & \includegraphics[width=0.09\linewidth]{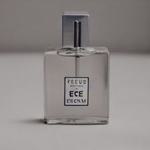} & 
\includegraphics[width=0.09\linewidth]{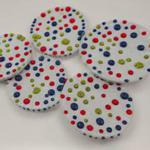} & \includegraphics[width=0.09\linewidth]{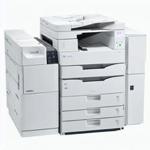} & \includegraphics[width=0.09\linewidth]{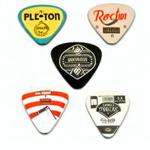} & \includegraphics[width=0.09\linewidth]{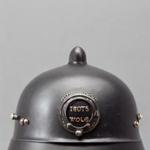} & \includegraphics[width=0.09\linewidth]{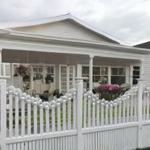} & \includegraphics[width=0.09\linewidth]{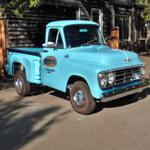} & \includegraphics[width=0.09\linewidth]{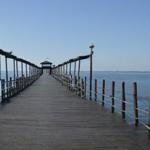} & \includegraphics[width=0.09\linewidth]{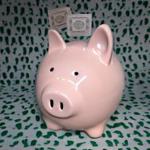} \\
\end{tabular}
\caption{More visualizations selected from the distilled ImageNet-1K (Class 700-799)}
\label{fig:vis8}
\end{figure*}

\clearpage
\begin{figure*}
\centering
\setlength{\tabcolsep}{0pt}
\renewcommand{\arraystretch}{0}
\begin{tabular}{llllllllll}
\includegraphics[width=0.09\linewidth]{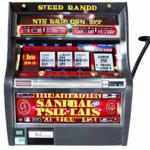} & \includegraphics[width=0.09\linewidth]{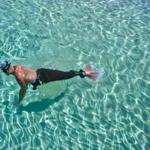} & \includegraphics[width=0.09\linewidth]{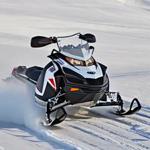} & \includegraphics[width=0.09\linewidth]{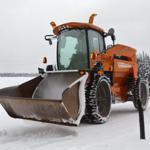} & \includegraphics[width=0.09\linewidth]{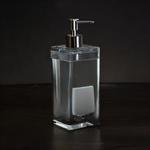} & \includegraphics[width=0.09\linewidth]{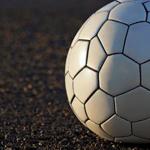} & \includegraphics[width=0.09\linewidth]{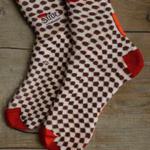} & \includegraphics[width=0.09\linewidth]{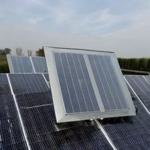} & \includegraphics[width=0.09\linewidth]{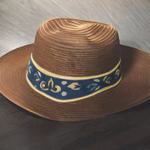} & \includegraphics[width=0.09\linewidth]{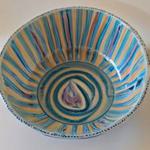} \\
\includegraphics[width=0.09\linewidth]{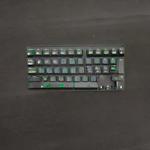} & \includegraphics[width=0.09\linewidth]{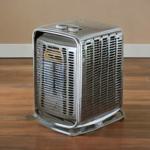} & \includegraphics[width=0.09\linewidth]{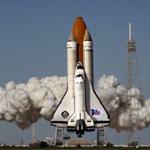} & \includegraphics[width=0.09\linewidth]{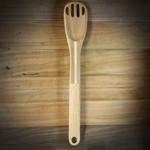} & \includegraphics[width=0.09\linewidth]{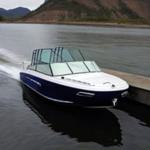} & \includegraphics[width=0.09\linewidth]{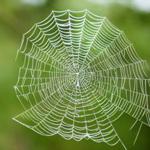} & \includegraphics[width=0.09\linewidth]{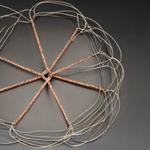} & \includegraphics[width=0.09\linewidth]{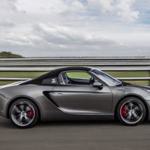} & \includegraphics[width=0.09\linewidth]{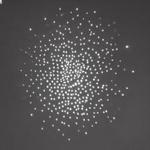} & \includegraphics[width=0.09\linewidth]{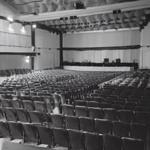} \\
\end{tabular}
\caption{More visualizations selected from the distilled ImageNet-1K (Class 800-899)}
\label{fig:vis9}
\end{figure*}

\clearpage
\begin{figure*}
\centering
\setlength{\tabcolsep}{0pt}
\renewcommand{\arraystretch}{0}
\begin{tabular}{llllllllll}
\includegraphics[width=0.09\linewidth]{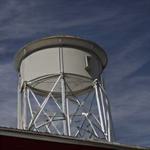} & \includegraphics[width=0.09\linewidth]{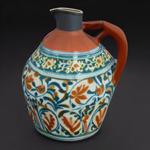} & 
\includegraphics[width=0.09\linewidth]{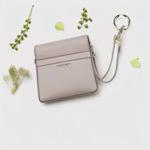} & \includegraphics[width=0.09\linewidth]{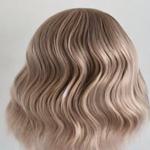} & \includegraphics[width=0.09\linewidth]{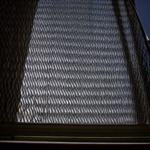} & \includegraphics[width=0.09\linewidth]{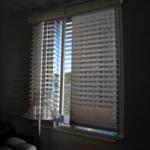} & \includegraphics[width=0.09\linewidth]{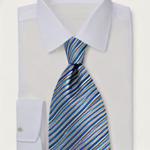} & \includegraphics[width=0.09\linewidth]{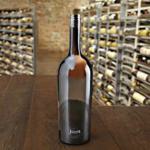} & \includegraphics[width=0.09\linewidth]{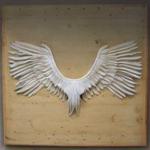} & \includegraphics[width=0.09\linewidth]{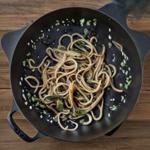} \\
\includegraphics[width=0.09\linewidth]{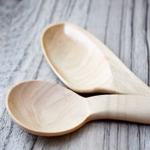} & \includegraphics[width=0.09\linewidth]{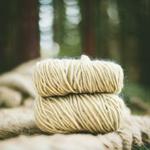} & 
\includegraphics[width=0.09\linewidth]{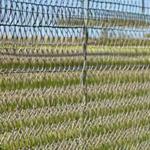} & \includegraphics[width=0.09\linewidth]{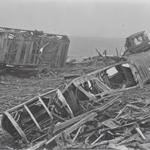} & \includegraphics[width=0.09\linewidth]{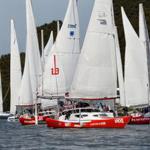} & \includegraphics[width=0.09\linewidth]{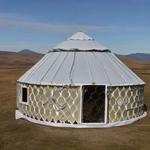} & \includegraphics[width=0.09\linewidth]{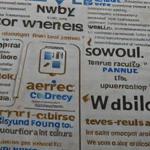} & \includegraphics[width=0.09\linewidth]{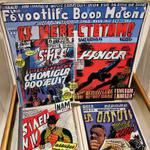} & \includegraphics[width=0.09\linewidth]{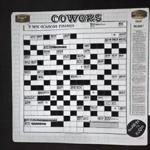} & \includegraphics[width=0.09\linewidth]{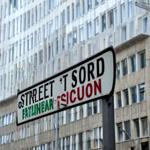} \\
\end{tabular}
\caption{More visualizations selected from the distilled ImageNet-1K (Class 900-999)}
\label{fig:vis10}
\end{figure*}